\documentclass[11pt]{article}

\newlength{\defbaselineskip}
\usepackage{algorithm}
\usepackage{algpseudocode}
\usepackage{framed}
\usepackage{amssymb}
\usepackage{amsfonts}
\usepackage{amsmath}
\usepackage{graphicx}
\usepackage{url}
\usepackage{rotating}
\usepackage{multirow}
\usepackage{color}
\usepackage{xcolor}
\usepackage{enumitem}
\usepackage{subfigure}

\usepackage{longtable} 
\usepackage{makecell}

\usepackage[multiple]{footmisc}
\usepackage[section]{placeins}

\usepackage{soul}

\usepackage{wrapfig}

\usepackage{hyperref}
\hypersetup{
     colorlinks   = true,
     linkcolor    = blue,
     citecolor    = green
}

\usepackage[normalem]{ulem}

\usepackage{xspace}
\newcommand{\SPECTRALNORM}{\texttt{LogSpectralNorm}\xspace}
\newcommand{\FROBENIUSNORM}{\texttt{LogFrobeniusNorm}\xspace}
\newcommand{\ALPHA}{\texttt{Alpha}\xspace}
\newcommand{\QUALITYOFALPHAFIT}{\texttt{QualityOfAlphaFit}\xspace}
\newcommand{\ALPHAHAT}{\texttt{AlphaHat}\xspace}
\newcommand{\ALPHASHATTENNORM}{\texttt{LogAlphaShattenNorm}\xspace}

\newcommand{\TASKONE}{\texttt{Task1}\xspace}
\newcommand{\TASKTWO}{\texttt{Task2}\xspace}
\newcommand{\BASELINE}{\texttt{Baseline}\xspace}

\newcommand{\AVGALPHA}{\langle\alpha\rangle}
\newcommand{\ALPHAHATEQN}{\hat\alpha}
\newcommand{\AVGALPHADISTANCE}{\langle D_{KS}\rangle}
\newcommand{\AVGLOGSPECTRALNORM}{\langle\log_{10}\Vert\mathbf{W}\Vert^{2}_{2}\rangle}
\newcommand{\AVGLOGNORM}{\langle\log_{10}\Vert\mathbf{W}\Vert^{2}_{F}\rangle}
\newcommand{\AVGLOGSHATTENNORM}{\langle\log_{10}\Vert\mathbf{W}\Vert^{2\alpha}_{2\alpha}\rangle}

\newcommand{\WW}{\texttt{WeightWatcher}}

\newcommand{\SHAPE}{\emph{Shape}\xspace}
\newcommand{\SCALE}{\emph{Scale}\xspace}

\begin{document}

\title{%
Post-mortem on a deep learning contest: a Simpson's paradox and the complementary roles of scale metrics versus shape metrics
}

\author{%
Charles H. Martin\thanks{Calculation Consulting, 8 Locksley Ave, 6B, San Francisco, CA 94122, \texttt{charles@CalculationConsulting.com}.} 
\and
Michael W. Mahoney\thanks{ICSI and Department of Statistics, University of California at Berkeley, Berkeley, CA 94720, \texttt{mmahoney@stat.berkeley.edu}.}
}

\date{}
\maketitle

\begin{abstract} 
To understand better good generalization performance in state-of-the-art neural network (NN) models, 
and in particular the success of the \ALPHAHAT metric based on Heavy-Tailed Self-Regularization (HT-SR) theory,
we analyze of a corpus of models that was made publicly-available for a 
contest to predict the generalization accuracy of NNs. 
These models include a wide range of qualities and were trained with a range of architectures and regularization hyperparameters.
We break \ALPHAHAT into its two subcomponent metrics: a scale-based metric; and a shape-based metric.
We identify what amounts to a Simpson's paradox: 
where ``scale'' metrics 
(from traditional statistical learning theory) 
perform well in aggregate, but can perform poorly on subpartitions of the data of a given depth,
when regularization hyperparameters are varied; and 
where ``shape'' metrics 
(from HT-SR theory) %
perform well on each subpartition of the data, 
when hyperparameters are varied for models of a given depth,
but can perform poorly overall when models with varying depths are aggregated.
Our results highlight 
the subtlety of comparing models when both architectures and hyperparameters are varied; 
the complementary role of implicit scale versus implicit shape parameters in understanding NN model quality; and
the need to go beyond one-size-fits-all metrics based on upper bounds from generalization theory to describe the performance of NN models. 
Our results also 
clarify further why the \ALPHAHAT metric from HT-SR theory works so well at predicting generalization across a broad range of CV and NLP models.

\end{abstract}

\vspace{-5mm}
\section{Introduction}
\label{sxn:intro}
\vspace{-1mm}

It is of increasing interest to develop metrics to measure the quality of Deep Neural Network (DNN) models, especially as models are applied well beyond their area of initial development.
This need is particularly acute when using publically-available, pre-trained models because one does not have regular access to training/testing data and/or information about the training protocols and hyperparameter choices.
Motivated by this, recent work introduced the \ALPHAHAT metric, (i.e $\hat{\alpha}$), and showed that this metric can predict trends in the quality, or generaliztion capacity, of state-of-the-art (SOTA) DNN models \emph{without access to any training or testing data}~\cite{MM20a_trends_NatComm}---outperforming other metrics from statistical learning theory (SLT) in a large meta-analysis of hundreds of SOTA models from computer vision (CV) and natural language processing (NLP).
The $\hat{\alpha}$ metric is based on the recently-developed Heavy-Tailed Self-Regularization (HT-SR) theory~\cite{MM18_TR_JMLRversion,MM19_HTSR_ICML,MM20_SDM}, which 
is based on statistical mechanics and Heavy-Tailed (HT) random matrix~theory.  

In this paper, we evaluate the \ALPHAHAT $(\hat{\alpha})$ metric (and it's subcomponents) on a series of pre-trained DNN models that were made publicly-available recently for a contest (``the Contest'') to predict generalization in deep learning~\cite{JFYx20_contest_v10,JFYx20_contest_v11}.
Recognizing problems with many metrics developed within traditional SLT, the Contest was interested in metrics that were ``causally informative of generalization,'' and it wanted participants to propose a ``robust and general complexity measure''~\cite{JFYx20_contest_v10,JFYx20_contest_v11}.
This set of Contest models was much more narrow than the corpus analyzed in the large-scale meta-analysis~\cite{MM20a_trends_NatComm}, e.g., the models were trained to only two CV tasks.
However, for that narrower class of models, the Contest data was more detailed, 
in the sense that there were lower quality models, models with a wider range of test accuracies, including models that generalize well, generalize poorly, and even models which appear to be overtrained.
Moreover, the models are partitioned into sub-groups of fixed depth, where only regularization hyperpaemeters are varied.
This more fine-grained set of pre-trained models provides us with the opportunity to evaluate the $\hat{\alpha}$ metric, and it's subcomponents,
across the opposing dimensions of depth and hyperparameter changes, more finely than was possible with the previous large-scale meta-analysis.

Our analysis of these pre-trained Contest models provides new insights---on how theories for generalization perform on well-trained versus poorly-trained models; 
on how this depends in subtle ways on what can be interpreted as implicit scale versus implicit shape parameters of the models learned by these DNNs; and
on how model quality metrics depend on architectural parameters versus solver parameters.
Most importantly, this work helps clarify why the \ALPHAHAT metric performs so well.

\paragraph{Background: Heavy-Tailed Self-Regularization (HT-SR) Theory.}
HT-SR theory is based on statistical mechanics%
\footnote{In the statistical mechanics approach to learning, generalization is related to volumes and thus the \SHAPE of a distribution, more than the 
\SCALE of the distribution~\cite{EB01_BOOK}.}
and HT random matrix theory; and it relies on the empirical fact that
for nearly all well-trained SOTA DNN models (in CV and NLP),
the individiual layer weight matrices $\mathbf{W}$ have
spectral densities $\rho_{}(\lambda)$ that are HT, and can be fit to a Power Law (PL) distribution~\cite{MM18_TR_JMLRversion}.%
\footnote{HT distributions are probability distributions with a relatively large amount of probability mass very far from the mean, i.e., in which ``unusual'' events are relatively likely to happen.  The precise functional form of the tail is less relevant for us, but for simplicity we will consider parameters from (truncated) PLs that have been fit to empirical data.  See Section~\ref{sxn:scale_shape_parameters} for details.}
Using layer PL fits, one can define the HT-based \ALPHAHAT metric for an entire DNN model.
This metric
can predict trends in the quality of SOTA DNNs, even without access to training or testing data~\cite{MM20a_trends_NatComm}.

Given a DNN weight matrix $\mathbf{W},\;(N\times M,\;N\ge M)$, let $\lambda$ be an eigenvalue of the correlation matrix $\mathbf{X}=\frac{1}{N}\mathbf{W}^{T}\mathbf{W}$.
The Empirical Spectral Density (ESD), $\rho_{emp}(\lambda)$, is the just an empirical 
fit of the histogram of the $M$ eigenvalues.
In looking at hundreds of models and thousands of weight matrices, the ESDs of well trained DNNs can nearly always be well fit to a PL distribution:
\vspace{-2mm}
\begin{equation}
\rho_{emp}(\lambda)\sim\lambda^{-\alpha},\;\; x_{min} \le \lambda \le x_{max}  .
\label{eqn:tpl_eqn}
\vspace{-2mm}
\end{equation}
Here, $x_{max}=\lambda_{max}$ is set to the  maximum eigenvalue of the ESD, and $x_{min}$ is the start of the tail of the ESD, fit as using the procedure of \cite{CSN09_powerlaw}.
The best fit is given by KS-distance $D_{KS}$ (denoted \QUALITYOFALPHAFIT below).
For models that generalize well, the fitted $\alpha\gtrsim 2.0$ for every (or nearly every)
layer \cite{MM18_TR_JMLRversion,MM20a_trends_NatComm}.

The \ALPHAHAT ($\hat{\alpha}$) metric from HT-SR theory
is a weighted average over $L$ layers of
two complementary subcomponents, 
the PL exponent $\alpha$ (\ALPHA), and the log maximum eigenvalue $\log\lambda^{max}$ (\SPECTRALNORM):
\vspace{-2mm}
\begin{equation}
\hat{\alpha}=\sum_{l=1}^{L}\alpha_{l}\log\lambda^{max}_{l}
\vspace{-2mm}
\end{equation}
This can be interpreted either as a weighted-average of \ALPHA values, weighted by the \SCALE
of the layer ESD $(\log\lambda^{max}_{l})$,
or as a weighted-average of the \SPECTRALNORM values, weighted by the \SHAPE of the layer ESD$(\alpha)$.
In this sense, it is useful to understand better how the $\hat{\alpha}$ metric works
by evaluating how each of these two subcompoents, the \ALPHA and \SPECTRALNORM metrics, 
individually perform when varying different opposing dimensions, such as the model depth (number of layers $L$)
and the model regularization (or solver) hyperameters $(\theta)$, such as dropout, momentum, weight decay, etc.

The Contest data allows us to do this.
However, in trying to understand the causes for \emph{why} \ALPHAHAT works,
we discovered that the component \ALPHA and \SPECTRALNORM metrics frequently display a \emph{Simpson's paradox}.%
\footnote{Simpson's paradoxes often arise in social science and biomedicine, since understanding causal relationships is particularly important in those areas. To our knowledge, this is the first application of these ideas in this~area.}
Being aware of challenges with designing good contests and with extracting causality from correlation, and knowing that one-size-fits-all metrics applied to heterogeneous data can lead to Simpson's paradoxes, we did not use the causal metrics provided by the Contest.
Instead, we adopted a different approach: we tried to identify and understand Simpson's paradoxes within the Contest data; and we used this to understand better the \ALPHAHAT and related generalization~metrics.

Recall that a \emph{Simpson's paradox} can arise when there is a larger data set that consists of multiple sub-groups; and
it refers to the situation where a predicted trend that holds for a larger data set consisting of multiple sub-groups disappears or reverses when those sub-groups are considered individually \cite{Sim51,BHO75,Rob09,KFWB13}.
Less relevant if one is simply interested in evaluating training/testing curves, or winning contests, 
Simpson's paradoxes are particularly relevant when one wants to understand the data, attribute causal interpretations to the underlying data/model, e.g., \emph{why} a particular metric performs well or poorly, or make decisions about how to improve a model, e.g., via architecture search, etc. 

Because \ALPHAHAT combines two complementary metrics, 
we consider two classes of metrics that have been used to predict DNN model quality: 
norm-based metrics from SLT (e.g., \SPECTRALNORM below); and 
HT-based metrics from HT-SR theory (e.g., \ALPHA below).
The norm-based metrics describe the \SCALE associated with the model implicitly-learned by the training process, 
and
have been used to provide generalization theory upper bound on simple models,%
\footnote{Basically, a bound on scale leads to a bound on generalization.}
but empirically they can perform in strange and counter-intuituve ways on even moderately-large realistic models~\cite{JNBx19_fantastic_TR}.
The HT (or other PL) metrics describe the \SHAPE of the implicitly-learned model,
are related to HT-SR theory~\cite{MM18_TR_JMLRversion,MM19_HTSR_ICML,MM20_SDM}, and have been 
conjectured to perform well when only varying solver hyperparmeters like batch-size.

\paragraph{Our main results.}

We apply these SLT and HT-SR metrics to predict generalization accuracies of the pre-trained models provided in the Contest, as a function of both model depth $L$ and regularization hyperparameters $\theta$.
Our main contributions are the following.

\textbf{New shape-based metrics.}
Based on our preliminary analysis of the data, we introduce two new metrics---\ALPHA and \QUALITYOFALPHAFIT---to evaluate model quality in the absence of training/testing data.
(Other metrics were used in previous work, but to our knowledge the use of these two metrics as model quality metrics is new.)

\textbf{Existence of Simpson's paradox in Contest data.}
We examine models provided by the Contest, and we identify Simpson's paradoxes.
Depending on the specific Contest task and model sub-group, \SPECTRALNORM and \ALPHA are either strongly anti-correlated or modestly to weakly anti-correlated with each other.
For both Contest tasks
(\TASKONE and \TASKTWO), 
and for each model sub-group, 
the fit trend 
in the \SPECTRALNORM,
at best, 
increases with increasing model quality (in \emph{dis}agreement with SLT).%
\footnote{Note that some fits, like \TASKONE 6xx, and \TASKTWO 2xx, are quite poor, further indicating the deficiencies in applying SLT to real world models.}
For \TASKONE, it increases when all models are considered together;
however, for \TASKTWO, it \emph{decreases} when all models are considered together, exhibiting a clear Simpson's paradox.%
\footnote{Here, the one possible exception is \TASKTWO 6xx, which we argue is an overtained model, as seen in Figure~\ref{fig:train-vs-test-task2}. }
Also, for both Contest tasks, and for each model subgroup, \ALPHA decreases with increasing model quality (in agreement with HT-SR theory).
For \TASKONE, it decreases when all models are considered together;
however, for \TASKTWO, it \emph{increases} when all models are considered together, again exhibiting a clear Simpson's paradox.

That is:
(1) for both \SPECTRALNORM and \ALPHA, the \TASKTWO models exhibit a Simpson's paradox;
(2) the predictions of \SPECTRALNORM \emph{disagree} with theory, except when aggregated when a Simpson's paradox is present; and 
(3) the predictions of \ALPHA \emph{agree} with theory, except when aggregated when a Simpson's paradox is~present.

\textbf{Shape parameters and hyperparameter variation.}
The \SCALE based \SPECTRALNORM can do well when the data are aggregated, but it does very poorly when the data are segmented by architecture type (in this case depth).
More generally, this metric is good at predicting test accuracy as large architectural changes, e.g., depth, are made.
However, it may behave in a manner opposite to that suggested by bounding theorems, meaning in particular that it can be \emph{anti-correlated} with test accuracies, when varying the regularization hyperparemeters
This confirms unexplained observations made in a different setting~\cite{JNBx19_fantastic_TR}.

The \SHAPE based \ALPHA from HT-SR theory is predictive of test accuracy, on a new/different dataset, as hyperparameters vary, when large-scale architectural changes are held fixed. 
Our results are the first to demonstrate that \ALPHA---a \SHAPE metric---is predictive of test accuracy, 
as model hyperparameters $\theta$ are varied independently.

Moreover, the \ALPHA metric performs better for higher quality models; this is evident
comparing results for the \TASKTWO 2xx versus 6xx models. 
For models with better test accuracies, the \ALPHA metric predicts them better.
This is seen visually in Figure~\ref{fig:simpsonsPlots2}, and also
in Table~\ref{table:quality_table_1}, which compares both linear and rank correlation metrics.

\textbf{Extracting scale and shape metrics from pre-trained DNN models.}
While computing norm-based \SCALE metrics is straightforward, computing HT-based \SHAPE metrics is much more subtle.
From HT-SR theory~\cite{MM18_TR_JMLRversion}, we want to fit the top part of the ESD of layer weight matrices to a PL distribution, as in Eqn.~(\ref{eqn:tpl_eqn}). 
This can be accomplished with the \texttt{WeightWatcher} tool)~\cite{weightwatcher_package}, as we describe below.
Since the parameters $x_{max}$ (the largest eigenvalue) and $\alpha$ (the PL exponent) in Eqn.~(\ref{eqn:tpl_eqn}) (below) have a natural interpretation in terms of the scale and shape of the PL distribution, we will interpret the corresponding \SPECTRALNORM and \ALPHA as empirically-determined (implicit) \SCALE and \SHAPE parameters for pre-trained DNN models.

\vspace{-2mm}
\section{Preliminaries and related~work}
\label{sxn:background}
\vspace{-1mm}

\paragraph{Predicting trends in model quality.}

There is a large body of (older~\cite{Bar97} and more recent~\cite{NTS15,BFT17_TR,AGNZ18_TR}) work from SLT on providing upper bounds on generalization quality of models; and there is a smaller body of (older~\cite{EB01_BOOK} and more recent~\cite{ZK16,MM18_TR_JMLRversion,BKPx20}) work using ideas from statistical mechanics to predict performance of models.
Most relevant for us are the recent results on ``Predicting trends in the quality of state-of-the-art neural networks without access to training or testing data''~\cite{MM20a_trends_NatComm} and ``Fantastic generalization measures and where to find them''~\cite{JNBx19_fantastic_TR}.  %
The former work~\cite{MM20a_trends_NatComm} considered a very broad range of CV and NLP models, including nearly every publicly-available pre-trained model, totaling to hundreds of SOTA models; and it focused on metrics that perform well on SOTA, without any access to training and/or testing data.
The latter work~\cite{JNBx19_fantastic_TR} considered a larger number models drawn from a much narrower range of CV models; 
and it considered a broad range of metrics, most of which require access to the training and/or testing~data.

There is also work that attempts to identify and evaluate empirically reliable generalization metrics, but using different evaluation measures than our simpler appoach of using just linear regression and rank correlation measures on different subgroups.
See, e.g.,~\cite{DDNx20_TR, LDRC18_TR, TPMx19_TR} and references therein.
In particular, the Contest used a casual measure, similar to that in \cite{JNBx19_fantastic_TR}, to evaluate all the models, for both tasks, \emph{in aggregate}.
We also point to~\cite{DDNx20_TR}, who used very different methods than ours to obtain conclusions consistent with ours.
They also noted that the kind of causal measures proposed by \cite{JNBx19_fantastic_TR} can obscure failures and successes; and
they argued that generalization studies should be evaluated using an aggregate measure of ``distributional robustness.''
However, they not did consider that Simpson's paradoxes can arise.
Indeed, causal relationships can not be infered directly from observational data \emph{because} of the presence of potential Simpson's paradoxes~\cite{Pearl09}.
We would not have been able to observe the tradeoff between \SCALE and \SHAPE metrics if we had chosen to use an aggregate evaluation measure.

Subsequent to the dissemination of the original technical report version of this article, similar shape-versus-scale ideas from HT-SR theory were used in a much more detailed analysis of NLP models~\cite{YTHx22_TR}.

\paragraph{Complexity metrics.}

For completeness, we have evaluated several quality metrics, including: 
two from SLT
(including \SPECTRALNORM and \FROBENIUSNORM)~\cite{JNBx19_fantastic_TR,JFYx20_contest_v10,JFYx20_contest_v11};
two from statistical mechanics and HT-SR Theory (including \ALPHAHAT and \ALPHASHATTENNORM)~\cite{MM18_TR_JMLRversion,MM19_HTSR_ICML,MM20_SDM}; and 
two (\ALPHA and \QUALITYOFALPHAFIT) that we introduce here.
See Table~\ref{table:metrics-considered} and the discussion in Appendix~\ref{app:metrics_in_analysis} for a summary of the best performing.  %
These metrics are implemented by the publicly-available \texttt{WeightWatcher} tool~\cite{weightwatcher_package}.
\paragraph{Models from a contest.}

The Contest~\cite{JFYx20_contest_v10,JFYx20_contest_v11} 
covered a rather narrow class of CV models, but it provided multiple versions of each, trained with different numbers of layers (depths) and regularization hyperparameters.
The approximately 150 CV models were organized into two architectural types, \emph{VGG-like models} and \emph{Network-in-Network models}, each with several sub-groups, 
trained with different batch sizes, dropout, and weight decay.  %

See Table~\ref{table:models-from-contest} 
in Appendix~\ref{app:models_in_analysis}
for a summary of the models and sub-groups.
There are two model groups: \TASKONE and \TASKTWO.
In every model sub-group (each row in Table~\ref{table:models-from-contest}), all models have the same depth (number of layers), 
and only solver hyperparameters are varied.
(In some cases, the weight matrices may have different shapes, e.g., $256 \times 256$ or $512 \times 512$.)
Briefly, the models are:
\begin{itemize}[leftmargin=*,noitemsep,nolistsep]
\item
\TASKONE (`task1\_v4''): $96$ VGG-like models, trained on CIFAR10, with 4 subgroups having 24 models each:  0xx, 1xx, 2xx, 5xx,  6xx, and 7xx.%
\footnote{\TASKONE also includes 2 different Convolutional widths for each, ${256, 512}$, and there is a large flattening layer which does not appear in actual VGG models.}
\item
\TASKTWO (``task2\_v1''): $54$ models, with stacked Dense layers, trained on SVHN, with 3 subgroups having 18 models each: 2xx, 6xx, 9xx, and 10xx.%
\footnote{The Dense layers in \TASKTWO are actually Conv2D($N,M,1,1$) layers, of width $512$}
\end{itemize}

\vspace{-2mm}
\section{Extracting scale and shape parameters from pre-trained models}
\label{sxn:scale_shape_parameters}
\vspace{-1mm}

Here, we describe how to identify scale versus shape metrics from DNN weight matrices, 
and we describe initial results comparing these metrics on the data from Table~\ref{table:models-from-contest}.

\vspace{-2mm}
\subsection{Computing scale and shape parameters with norms and PL exponents}
\label{sxn:scale_shape_parameters-overview}
\vspace{-1mm}

Here, we describe the metrics we consider.
See Appendix~\ref{app:metrics_in_analysis} for a more precise definition and detailed discussion.

\vspace{-2mm}
\paragraph{``Scale'' parameters.}

The \SPECTRALNORM and \FROBENIUSNORM are the average over all layer weight matrices of the logarithm of the corresponding norm~\cite{JNBx19_fantastic_TR}.
Since weight matrices from Table~\ref{table:models-from-contest} are not extremely large, computing \SPECTRALNORM 
(e.g., with the power method) 
and \FROBENIUSNORM 
(e.g., by summing the squares of all the entries) 
is simple. 
These are norms, 
and each has a natural interpretation as a ``scale'' parameter for the model implicitly learned by the NN.%
\footnote{Clearly, this metric ignores inter-layer structure, etc.  In light of our results, developing improved methods to extract better location/scale/shape parameters from pre-trained DNN models is an important direction for future~work.}

\vspace{-2mm}
\paragraph{``Shape'' parameters.}

To extract finer-scale ``shape'' information, recall that the spectral/Frobenius norms are unitarily invariant, i.e., they depend just on the vector of matrix singular values. 
We can remove the \SCALE of this vector by normalizing it to be a probability distribution; and then we can fit the \SHAPE of this distribution to a PL distribution (or some other HT distribution).

The \ALPHA metric reported here is the \emph{average PL exponent} over all DNN layers,
fit using the publically available and open-source \WW tool~\cite{weightwatcher_package}.  
Fitting HT distributions has subtleties~\cite{CSN09_powerlaw,newman2005_zipf},
and \WW~encaspulates the process, making it straightforward and reproducible.%
\footnote{Note that we do \emph{not} include outliers (defined as when $\alpha\ge 8$) in the layer average.
This indicates that either this layer has a low-quality PL fit, or that the layer weight matrix is not HT but instead more Random-Like~\cite{MM18_TR_JMLRversion}.}
The \QUALITYOFALPHAFIT metric is also a layer average, determined using a Kolmogorov-Smirnov Goodness of Fit Test (or KS-test), denoted $D_{KS}$.
Based on this, we can interpret the parameters of this fit---$\alpha$ and $x_{max}$---as \SHAPE and \SCALE parameters, respectively.
Neither \ALPHA nor \QUALITYOFALPHAFIT have been considered previously, i.e., both are new to our~work.

\paragraph{Combining shape and scale.}
We also evaluate metrics from previous work that we can now interpret as combining shape and scape:
\ALPHAHAT~\cite{MM20a_trends_NatComm}
and  
\ALPHASHATTENNORM~\cite{MM20_SDM}.   

\vspace{-2mm}
\subsection{Fitting ESDs to PLs}
\label{sxn:scale_shape_parameters-fitting}
\vspace{-1mm}

In practice, these PL fits require some care to obtain consistent, reliable results; and, for this reason, here we describe some of these issues.
Among other things, it is
important to understand the behavior of the PL fit as $x_{min}$ is varied.
See Figure~\ref{fig:good-vs-bad-pl-fits-ideal} for an illustrative (very ``nice'') example; and
see Appendix~\ref{app:fitting_tpls} for more discussion of these fitting issues, including dealing with non-ideal~ESDs.
(These figures can be genertated by the \WW~tool.)
\begin{figure}[t] %
    \centering
    \subfigure[ESD (log-log plot)]{
        \includegraphics[width=3.7cm]{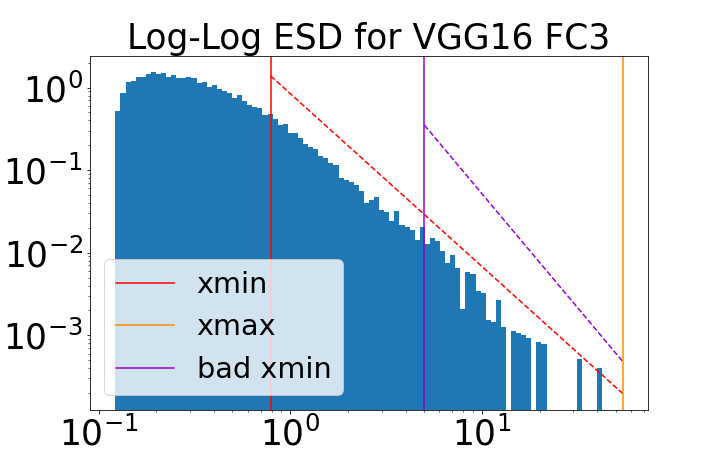}
        \label{fig:good-vs-bad-pl-fits-ideal-11}
    }
    \subfigure[ESD (lin-lin plot)]{
        \includegraphics[width=3.7cm]{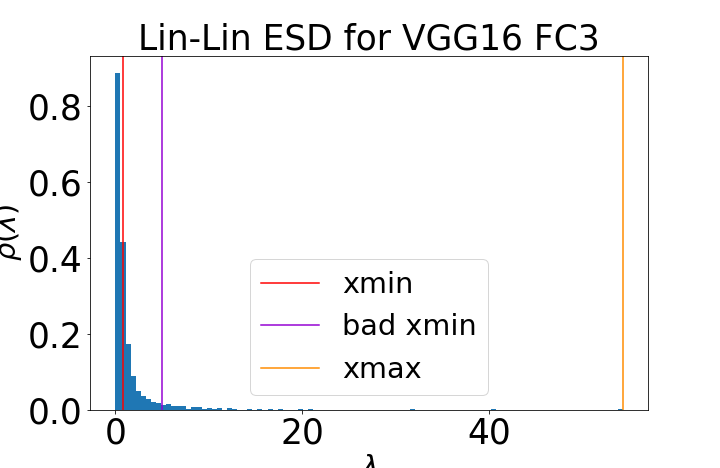}
        \label{fig:good-vs-bad-pl-fits-ideal-12}
    }
    \subfigure[ESD (log-lin plot)]{
        \includegraphics[width=3.7cm]{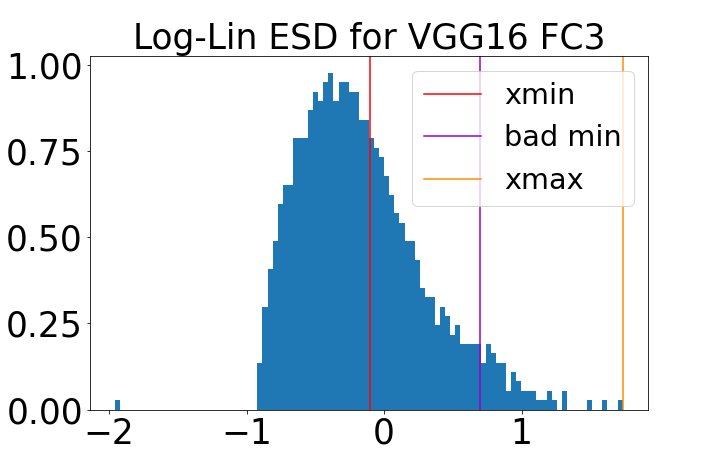}
        \label{fig:good-vs-bad-pl-fits-ideal-13}
    }
    \subfigure[PL fit quality vs $x_{min}$]{
        \includegraphics[width=3.7cm]{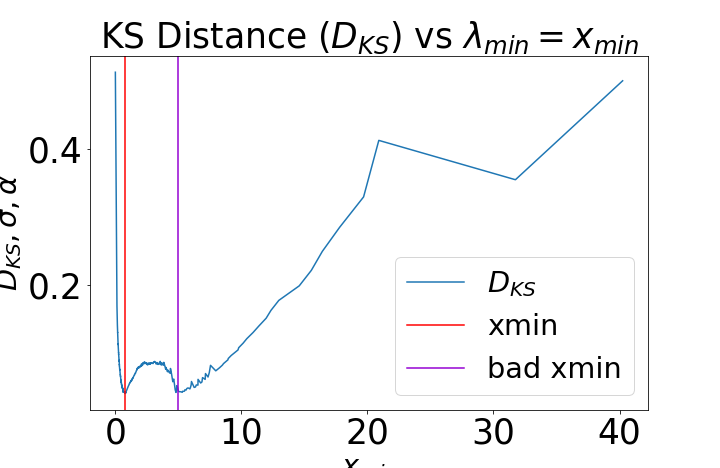}
        \label{fig:good-vs-bad-pl-fits-ideal-14}
    } 
    \caption{Illustration of the role of the ESD shape in determining the PL parameter $\alpha$ in the PL fit.  
             Shown is VGG16, FC3 (a nearly ``ideal'' example).
             (See Appendix~\ref{app:fitting_tpls} for less ideal~examples.)
            }
    \label{fig:good-vs-bad-pl-fits-ideal}
\end{figure}

In 
Figure~\ref{fig:good-vs-bad-pl-fits-ideal}, we consider the third fully connected (FC) layer from VGG16 (which was studied previously~\cite{MM18_TR_JMLRversion,MM20a_trends_NatComm}), which is a nearly ``ideal'' example that we use to illustrate the method.  %
In~\ref{fig:good-vs-bad-pl-fits-ideal-11}, \ref{fig:good-vs-bad-pl-fits-ideal-12}, and~\ref{fig:good-vs-bad-pl-fits-ideal-13}, respectively, we show the ESD in log-log plot, linear-linear plot, and log-linear plot; and in each of those plots, we mark the position of $x_{max}$, the $x_{min}$ found by the fitting procedure, and a suboptimal value of $x_{min}$ chosen by hand.

The log-log plots, in~\ref{fig:good-vs-bad-pl-fits-ideal-11} highlight the (well-known but often finicky) linear trend on a log-log plot.
To aid the eye, in~\ref{fig:good-vs-bad-pl-fits-ideal-11}, we show the slopes on a log-log plot, as determined by the $\alpha$ fit for that value of $x_{min}$.
The linear-linear plots, in~\ref{fig:good-vs-bad-pl-fits-ideal-12}, the usual so-called scree plots, are not particularly informative.
The log-linear plots, in~\ref{fig:good-vs-bad-pl-fits-ideal-13}, shows that the optimal value of $x_{min}$ is near the peak of the distribution, and that the distribution is clearly not log-normal, having a strong right-ward bias, with a spreading out of larger eigenvalues as $\lambda_{max}$ is~approached.

In~\ref{fig:good-vs-bad-pl-fits-ideal-14}, we show the quality of the fit, measured by the KS distance, as a function of $x_{min}$.
The interpretation of these PL fits is that, above that value of $x_{min}$ and below the value of $x_{max}$, the ESD is fit to a line, with slope $-\alpha$, in a log-log plot.
In~\ref{fig:good-vs-bad-pl-fits-ideal-14}, we see that the suboptimal value of $x_{min}$ was chosen to be a local minimum in the the KS distance plot, and that the fit quality gradually degrades as $x_{min}$ increases.
Looking at~\ref{fig:good-vs-bad-pl-fits-ideal-13}, we see that there is a very slight ``shelf'' in the ESD probability mass; and that the suboptimal value of $x_{min}$ corresponds to fitting a much smaller portion of the ESD with a PL fit.%
\footnote{Indeed, choosing $x_{min}$ introduces a scale, somewhat like fixing a rank parameter---since where the PL exponent starts is related to the dimensionality of the subspace of the low-rank approximation.  As opposed to more popular rank parameter selection methods, choosing $x_{min}$ simply from visual inspection of the ESDs, without computing the KS distance as a function of $x_{min}$, seems not to be possible when working with PL distributions.}
For high-quality models, like this one, smaller values of $\alpha$ corresponds to better models, and decreasing $\alpha$ is well-correlated with increasing $\lambda_{max}$.%
\footnote{This is predicted by HT-SR Theory~\cite{MM18_TR_JMLRversion,MM20a_trends_NatComm}, but it can be surprising from the perspective of SLT~\cite{JNBx19_fantastic_TR,JFYx20_contest_v10,JFYx20_contest_v11}.}

\vspace{-2mm}
\subsection{Comparing scale versus shape parameters}
\label{sxn:scale_shape_parameters-comparing}
\vspace{-1mm}

\begin{figure}[t]  %
    \centering
    \subfigure[\TASKONE models.]{
        \includegraphics[width=4.9cm]{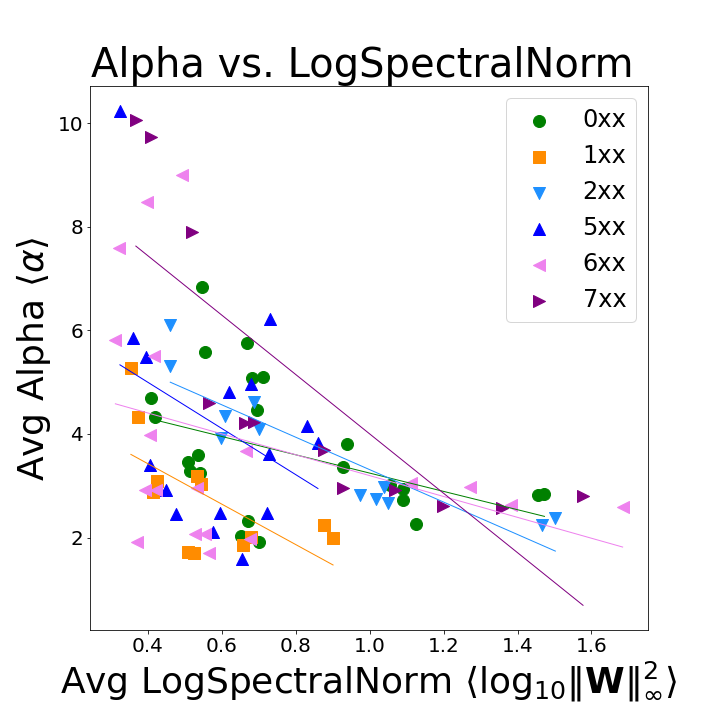}
        \label{fig:alpha-versus-spectral-A}
    }
    \subfigure[\TASKTWO models.]{
        \includegraphics[width=4.9cm]{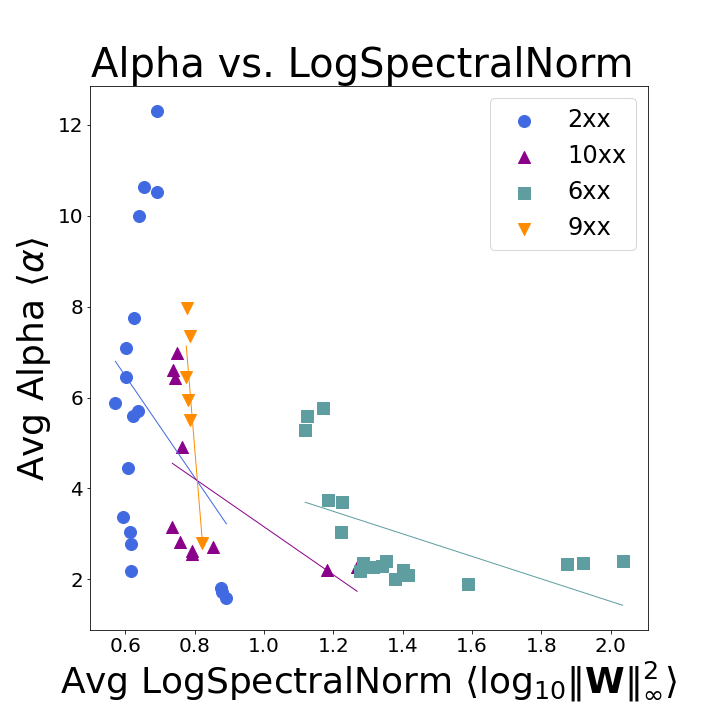}
        \label{fig:alpha-versus-spectral-B}
    }
    \caption{Comparison of the \SPECTRALNORM and \ALPHA metrics, for \TASKONE and \TASKTWO models.
             Observe that, depending on the subtask, these implicitly-defined model ``scale'' and ``shape'' parameters can behave very differently.
            }
    \label{fig:alpha-versus-spectral}
\end{figure}

Here, we compare \SHAPE versus \SCALE parameters, illustrating that they capture different information about models, as task, depth, and regularization / solver hyperparameters are varied.
See Figure~\ref{fig:alpha-versus-spectral}, which compares \ALPHA and \SPECTRALNORM for models from Table~\ref{table:models-from-contest}, segmented into sub-groups corresponding to models with the same depth.
See also Table~\ref{table:alpha-vs-spnorm} for more detailed numerical results. 
For some model sub-groups, \ALPHA and \SPECTRALNORM are strongly anti-correlated; while for other sub-groups, they are modestly to weakly anti-correlated, at best.

For example, for \TASKONE, 
in the $2xx$ and $9xx$ models, the two metrics are strongly anti-correlated metrics, with $R^2>0.6$ and a Kendall-$\tau$ rank correlation metric of $\tau>0.75$;%
\footnote{A strong linear correlation, not just good rank correlation, is predicted for good models by HT-SR theory~\cite{MM18_TR_JMLRversion}, and this has been observed previously for SOTA CV and NLP models~\cite{MM20a_trends_NatComm}.}
in the $1xx$ models, they are modestly so, with $R^2\sim0.4$ and $\tau\sim0.4$; and  
in the remaning model sub-groups, $R^2<0.2$ and $\tau\le0.3$, which we identify as weakly correlated.
Similarly, in \TASKTWO, we can identify 
model sub-group $9xx$ has having well anti-correlated average metrics, with large $R^2$ and Kendall-$\tau$;
model sub-groups $6xx$ and $10xx$ exhibit intermediate behavior, as $R^2<0.3$ for each; and 
model sub-group $2xx$ shows no substantial correlation at all between \ALPHA and \SPECTRALNORM.
See Appendix~\ref{app:illustrative_scale_shape} for more details on two illustrative pairs of examples.

\begin{table}[t] %
\small
\begin{center}
\begin{tabular}{|p{1.00in}|c|c|c|}
\hline
               &  $R^2$ & Kendall-$\tau$ & Correlation \\
\hline
\TASKONE - 0xx & 0.162 & 0.29  & Weak   \\
\TASKONE - 1xx & 0.405 & 0.394 & Modest \\
\TASKONE - 2xx & 0.803 & 0.788 & Strong \\
\TASKONE - 5xx & 0.124 & 0.117 & Weak   \\
\TASKONE - 6xx & 0.124 & 0.263 & Weak   \\
\TASKONE - 7xx & 0.64 & 0.909  & Strong \\
\hline
\TASKONE - AVG & 0.38 & 0.46 & \\
\hline
\hline
\TASKTWO - 2xx  & 0.113 & 0.0327 & None   \\
\TASKTWO - 6xx  & 0.282 & 0.451  & Modest \\
\TASKTWO - 9xx  & 0.754 & 0.600  & Strong \\
\TASKTWO - 10xx & 0.273 & 0.636  & Modest \\
\hline
\TASKTWO - AVG & 0.36 & 0.43 & \\
\hline
\end{tabular}
\end{center}
\caption{Comparison of the \ALPHA and \SPECTRALNORM metrics, for \TASKONE and \TASKTWO models.
         $R^2$ and Kendall-$\tau$ for \TASKONE and \TASKTWO, both aggregated and partitioned into model sub-groups.
}
\label{table:alpha-vs-spnorm} 
\end{table}

\vspace{-3mm}
\section{A Simpson's paradox: Architecture versus solver knobs}
\label{sxn:simpsons_paradox}
\vspace{-1mm}

\subsection{Basic properties of the data}
\label{sxn:simpsons_paradox-basic}
\vspace{-1mm}

\begin{figure}[t] %
    \centering
    \subfigure[\TASKONE models.]{
        \includegraphics[width=5cm]{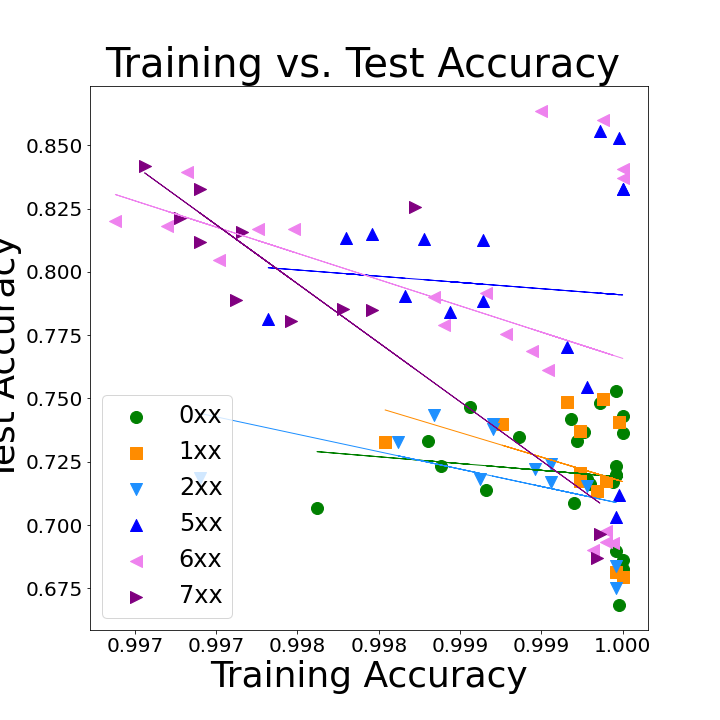}
        \label{fig:train-vs-test-task1}
    }
    \subfigure[\TASKTWO models.]{
        \includegraphics[width=5cm]{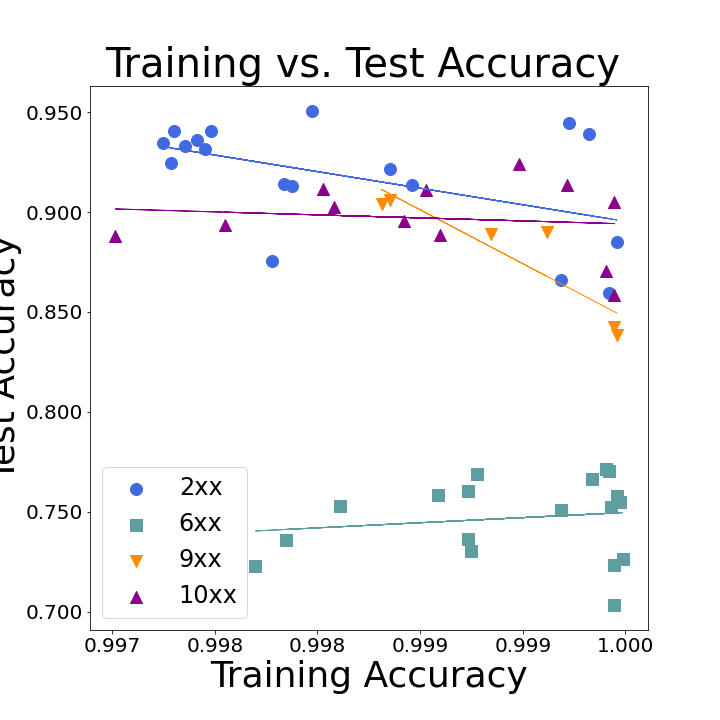}
        \label{fig:train-vs-test-task2}
    }
    \caption{Relationship between training accuracy and testing accuracy for \TASKONE and \TASKTWO~models.
             One would expect a positive correlation or (if the training error is very close to zero) at least not a negative correlation.
             Observe, however, that in many cases they are strongly anti-correlated.
             See also Table~\ref{table:trainingVersusTest}.
            }
    \label{fig:trainingVersusTest}
\end{figure}

The obvious \BASELINE for model quality is the training accuracy.
If we have access only to pre-trained models and no data (as in prior work~\cite{MM20a_trends_NatComm} and as is assumed with \ALPHA and \SPECTRALNORM), then we cannot check this baseline.
Similarly, if the training error is \emph{exactly} zero, then this is not a useful baseline.%
\footnote{Models that \emph{exactly} interpolate the data are different, in that we can't just ``add data'' (and without retraining work with the same baseline), since then we won't have exactly $0\%$ error with the augmented data set.  For many model diagnostics, exactly interpolating and approximately interpolating are \emph{very} different.}
Otherwise, we can check against it, and we expect testing accuracy to improve as training accuracy~improves.

In a practical setting, for a properly trained model, one expects that test accuracy will increase with increasing training accuracy, or at least not decrease.  
If the test accuracy does decrease, this indicates that the model is somewhat overtrained and therefore will not generalize well.
Figure~\ref{fig:trainingVersusTest} plots the relationship between training and testing accuracies for the \TASKONE and \TASKTWO models, color-coded by model\_number (0xx, 1xx, \ldots). 
See also Table~\ref{table:trainingVersusTest} for more detailed numerical results.
By looking at the both the trendlines and the distrbution of points in Figure~\ref{fig:trainingVersusTest}, we can identify such overtrained models here: almost all of the \TASKONE models (except perhaps the very highly accurate 5xx and 6xx models, and the moderately accurate 0xx and 1xx models); the \TASKTWO 9xx model; and to a lesser extent the \TASKTWO 2xx model.

For both \TASKONE and \TASKTWO, there is a very large gap between the training and testing accuracies.
A positive correlation in Figure~\ref{fig:trainingVersusTest} indicates that improving the training accuracy, even marginally, would lead to improved testing accuracy.
Instead, we see that for \TASKONE, for most model sub-groups (1xx, 2xx, 6xx, and 7xx), training and testing accuracies are (very) anti-correlated.
For other two (0xx and 5xx), the trendline is only weakly anti-correlated. For all models, however,
as the training accuracy approaches 1, the test accuracy plummets (except for a few outliers in the 5xx and 6xx groups).
In this sense, these \TASKONE models are over-trained; they clump into two sub-groups, those for this $R^2 \approx 0$ and those for which $R^2 > 0.1$.
Similarly, for \TASKTWO, for model\_number 9xx, they are (very) anti-correlated, and otherwise they are (slightly) anti-correlated for 2xx.
The other two models show essentially no discernable trend.%
\footnote{In a Contest, one would expect test data to reflect accurately the qualitative nature of the (secret) Contest data; these results suggest that this may not be the case.}
In all cases, improving training accuracy leads either to no noticible improvement in test accuracy, or to worse testing~performance.

Overall, many of the \TASKONE models, as well as model \TASKTWO 9xx, behave both qualitatively and quantitively differently than the reasonably well-trained \TASKTWO models.
We believe this is because they may be overtrained.

\begin{table}[t] %
\small
\begin{center}
\begin{tabular}{|p{0.90in}|c|c|c|c|}
\hline
{} &   $R^2$ &  RMSE &  Kendall-$\tau$ & Correlation\\
\hline
\TASKONE 0xx  & 0.01 &  0.02 &  0.10 & Weak \\
\TASKONE 1xx  & 0.12 &  0.02 &  0.33 & Modest  \\
\TASKONE 2xx  & 0.22 &  0.02 &  0.54 & Strong \\
\TASKONE 5xx  & 0.01 &  0.04 &  0.01 & Weak \\
\TASKONE 6xx  & 0.16 &  0.05 &  0.31 & Modest \\
\TASKONE 7xx  & 0.78 &  0.02 &  0.60 & Strong \\
\hline
\TASKONE AVG  & 0.22 &  0.03 &  0.31 & \\
\hline
\hline
\TASKTWO 2xx  & 0.24 &  0.02 &  0.25 & Modest \\
\TASKTWO 6xx  & 0.02 &  0.02 & -0.05 & Weak \\
\TASKTWO 9xx  & 0.83 &  0.01 &  0.73 & Strong  \\
\TASKTWO 10xx & 0.02 &  0.02 &  0.02 & Weak \\
\hline
\TASKTWO AVG  & 0.28 &  0.02 &  0.24 & \\
\hline
\end{tabular}

\end{center}
\caption{Quality metrics 
        (for $R^2$, larger is better; for RMSE, smaller is better; and for Kendall-$\tau$ rank correlation, larger magnitude is better) 
        for the relationship between training error and testing error, as illustrated in Figure~\ref{fig:trainingVersusTest}.
}
\label{table:trainingVersusTest}
\end{table}

\vspace{-2mm}
\subsection{Visualizing the Simpson's paradox in DNN models}
\label{sxn:simpsons_paradox-visualized}
\vspace{-1mm}

\begin{figure}[t] %
    \centering
    \subfigure[\TASKONE models.]{
        \includegraphics[width=5cm]{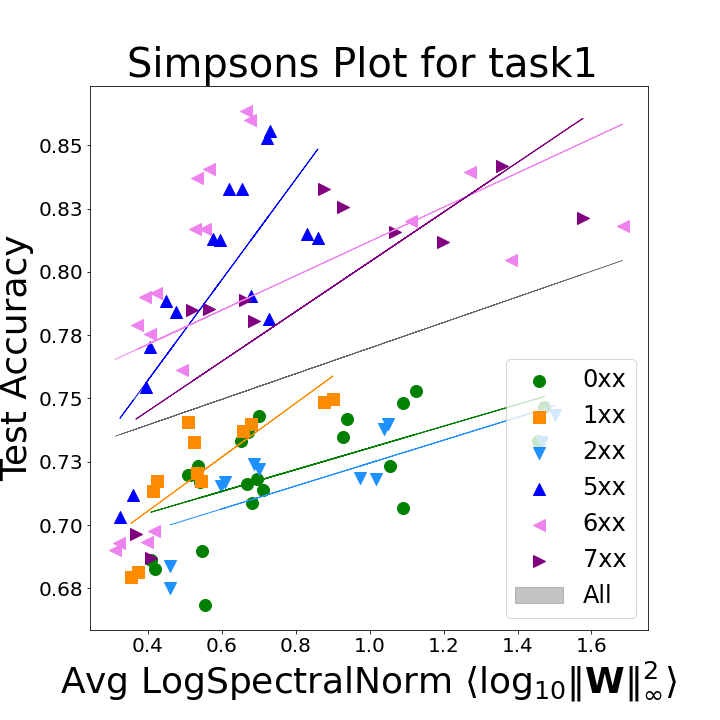}
        \label{fig:simpsonsPlots1-task1}
    }
    \subfigure[\TASKTWO models.]{
        \includegraphics[width=5cm]{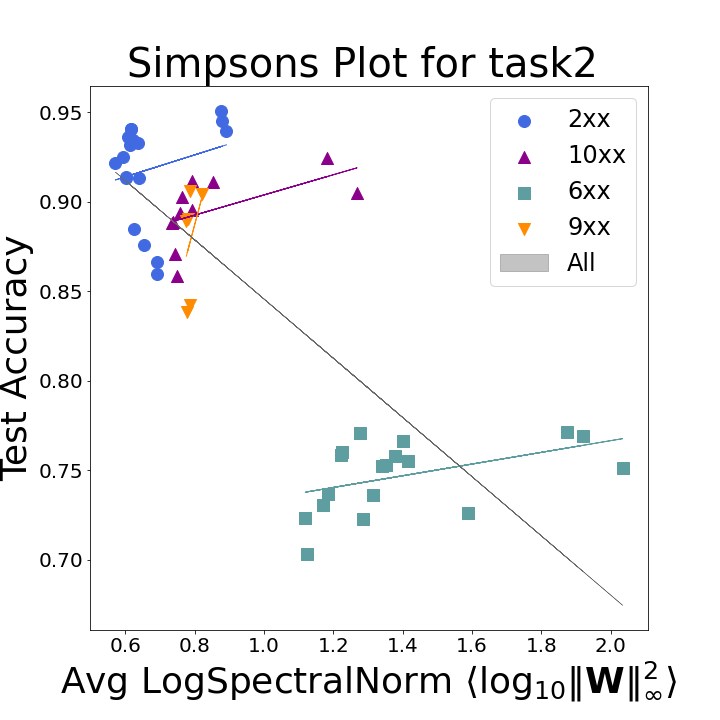}
        \label{fig:simpsonsPlots1-task2}
    }
    \caption{Test accuracy versus \SPECTRALNORM, for \TASKONE and \TASKTWO, overall and segmented by model sub-group.
             Observe the clear Simpson's paradox, in particular for \TASKTWO models.  }
    \label{fig:simpsonsPlots1}
\end{figure}

\begin{figure}[t] %
    \centering
    \subfigure[\TASKONE models.]{
        \includegraphics[width=5cm]{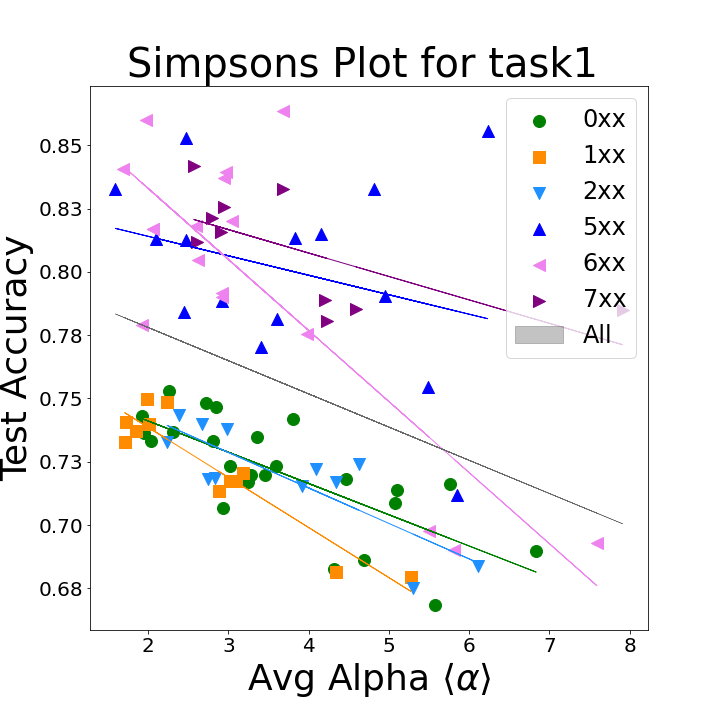}
        \label{fig:simpsonsPlots2-task1}
    }
    \subfigure[\TASKTWO models.]{
        \includegraphics[width=5cm]{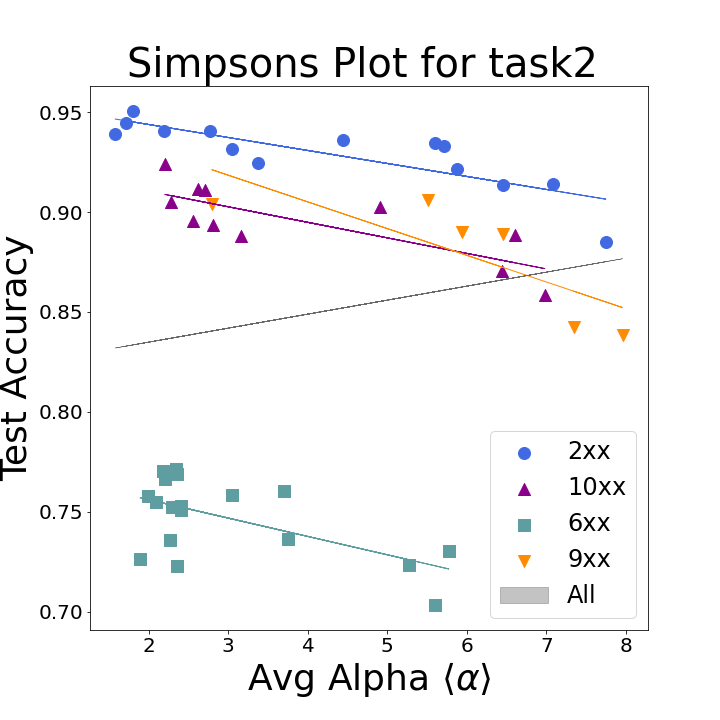}
        \label{fig:simpsonsPlots2-task2}
    }
    \caption{Test accuracy versus \ALPHA, for \TASKONE and \TASKTWO, overall and segmented by model sub-group.
             Observe the clear Simpson's paradox, in particular for \TASKTWO models.  
    }
    \label{fig:simpsonsPlots2}
\end{figure}

We now consider how the 
test accuracy varies against the \SPECTRALNORM and \ALPHA, respectively, overall and broken down for each model sub-group, for all models from Table~\ref{table:models-from-contest}.
See Figure~\ref{fig:simpsonsPlots1} and Figure~\ref{fig:simpsonsPlots2} for a summary.
In our analysis, we first look at each model sub-group (0xx, 1xx, 2xx, \ldots) individually, which corresponds to a specific depth $L$, and we measure the regression and rank correlation metrics on the test accuracy as the hyperparemeters vary.

From Figure~\ref{fig:simpsonsPlots1}, for each model sub-group, the \SPECTRALNORM increases with increasing model quality.
For \TASKONE, the \SPECTRALNORM increases when all models are considered together; but 
for \TASKTWO, the \SPECTRALNORM \emph{decreases} when all models are considered together, exhibiting a clear Simpson's paradox.

From Figure~\ref{fig:simpsonsPlots2}, for each model subgroup, the \ALPHA decreases with increasing model quality.
For \TASKONE, the \ALPHA decreases when all models are considered together; but 
for \TASKTWO, the \ALPHA \emph{increases} when all models are considered together, exhibiting a clear Simpson's~paradox.

Both metrics exhibit a 
clear 
Simpson's paradox for \TASKTWO.

Based on bounding theorems from statistical learning theory, one might expect that smaller values of \SPECTRALNORM would correspond to better models.
Similarly, based on HT-SR theory, one would expect that smaller values of \ALPHA would correspond to better models.
With respect to these references, the \SPECTRALNORM behaves the \emph{opposite of what theory would suggest, except when} considering the aggregated data when a Simpson's paradox is present (i.e., in \TASKTWO).
On the other hand, the \ALPHA behaves \emph{precisely as what theory would predict, except when} considering the aggregated data when a Simpson's paradox is present.

\vspace{-2mm}
\subsection{Changing architectures versus changing solver hyperparameters}
\label{sxn:simpsons_paradox-details}
\vspace{-1mm}

\begin{figure}[t] %
    \centering
    \subfigure[\TASKONE, $R^2$]{
        \includegraphics[width=3.7cm]{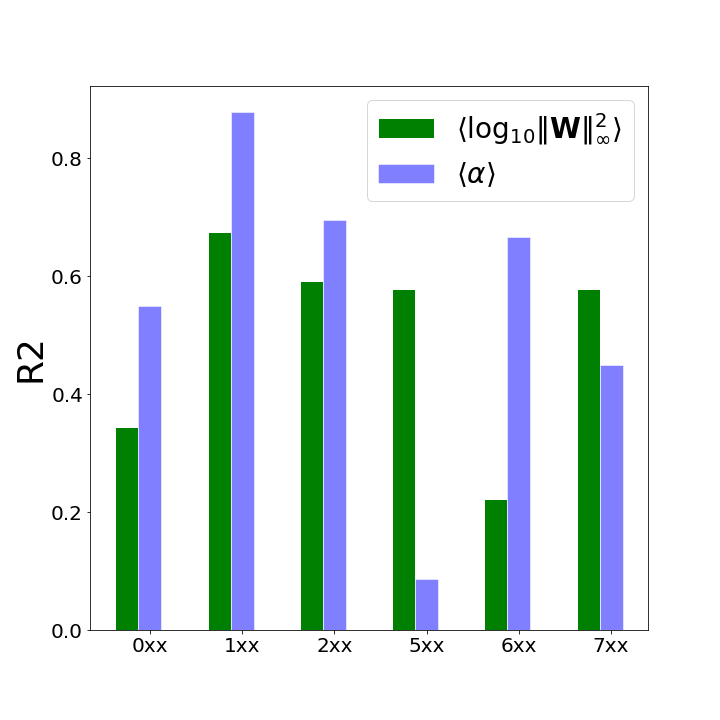}
        \label{fig:task1-r2-test-accs}
    }
    \subfigure[\TASKTWO, $R^2$]{
        \includegraphics[width=3.7cm]{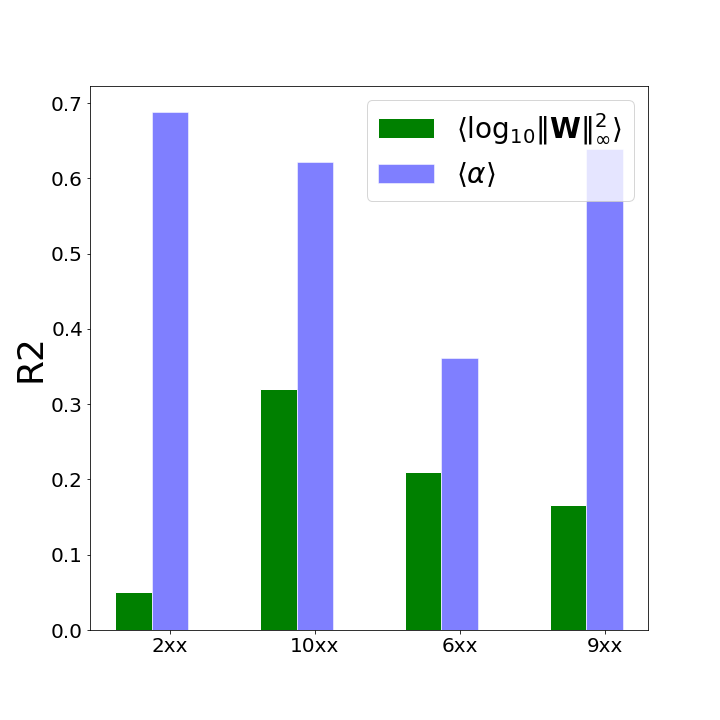}
        \label{fig:task2-r2-test-accs}
    } 
    \subfigure[\TASKONE, Kendall-$\tau$]{
        \includegraphics[width=3.7cm]{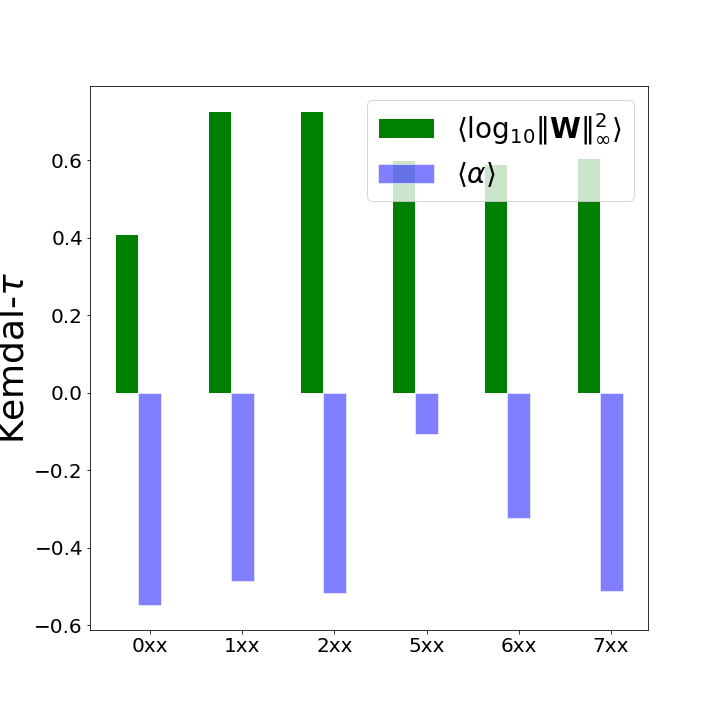}
        \label{fig:task1-ktau-test-accs}
    }
    \subfigure[\TASKTWO, Kendall-$\tau$]{
        \includegraphics[width=3.7cm]{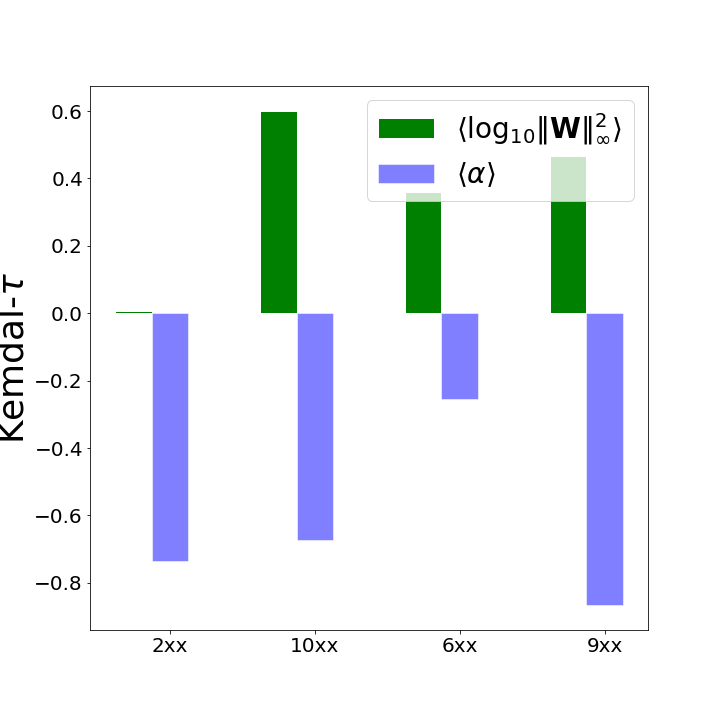}
        \label{fig:task2-Ktau-test-accs}
    }
    \caption{Comparison of predictions for test accuracies by \ALPHA $\AVGALPHA$ and \SPECTRALNORM $\AVGLOGSPECTRALNORM$, as measured using $R^2$ and Kendall-$\tau$ correlation and rank correlation metrics, respectively, for \TASKONE and \TASKTWO, for each model sub-group.
             (See Table~\ref{table:quality_table_1} for more details.)
            }
    \label{fig:alpha-vs-spnorm-test-accs}
\end{figure}

Here, we provide a more detailed analysis of the main results presented in Figure~\ref{fig:simpsonsPlots1} and Figure~\ref{fig:simpsonsPlots2}.
See Figure~\ref{fig:alpha-vs-spnorm-test-accs} for histograms summarizing some of these results, in particular for \ALPHA and \SPECTRALNORM.
See also Table~\ref{table:quality_table_1} and Table~\ref{table:quality_table_2} 
in Appendix~\ref{app:details-simpsons-changing}
for detailed statistics, including $R^2$ and Kendall-$\tau$ statistics, for metrics from Table~\ref{table:metrics-considered}.

From these, we see that
\ALPHA (the mean PL exponent, averaged over all layers, which corresponds to a \SHAPE parameter)
\emph{is correlated with the test accuracy for each DNN, when changing just the hyperparameters, $\theta$}.
While it does not always exhibit the strongest correlation, it is the most consistent.
We also see that
\SPECTRALNORM (the mean $\log_{10}$ spectral norm, also averaged over all layers, which corresponds to a \SCALE parameter)
\emph{is correlated with the test accuracy, when changing the number of layers, but it is anti-correlated when changing the hyperparameters}.
Thus, it performs quite poorly when trying to identify more fine-scale structure. 

For completeness, we have included \FROBENIUSNORM, showing that it is often but not always correlated with test accuracy, and \QUALITYOFALPHAFIT (the KS distance), showing that it is correlated with test accuracy in about half the cases.
\begin{figure}[t] %
    \centering
    \subfigure[\TASKONE.]{
        \includegraphics[width=5.0cm]{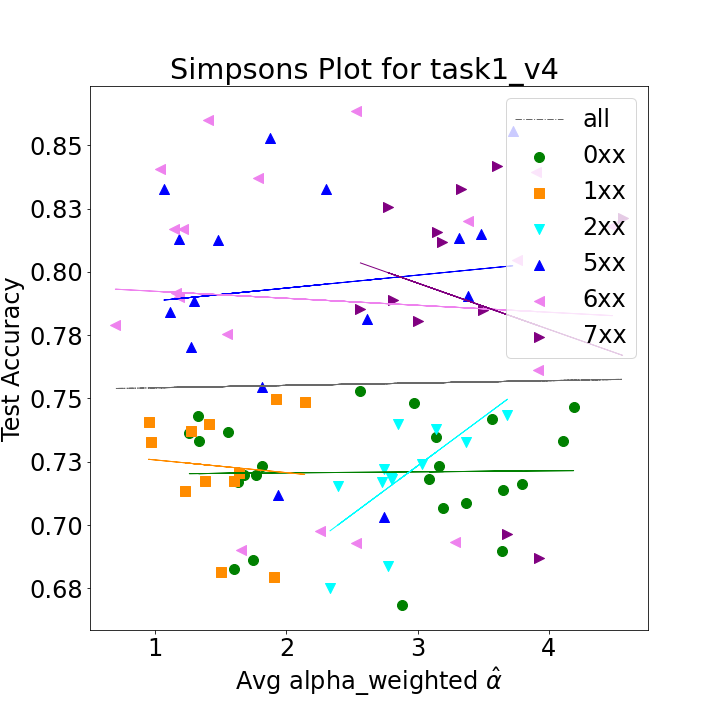}
        \label{fig:simpsonsPlots3-task1}
    }
    \subfigure[\TASKTWO.]{
        \includegraphics[width=5.0cm]{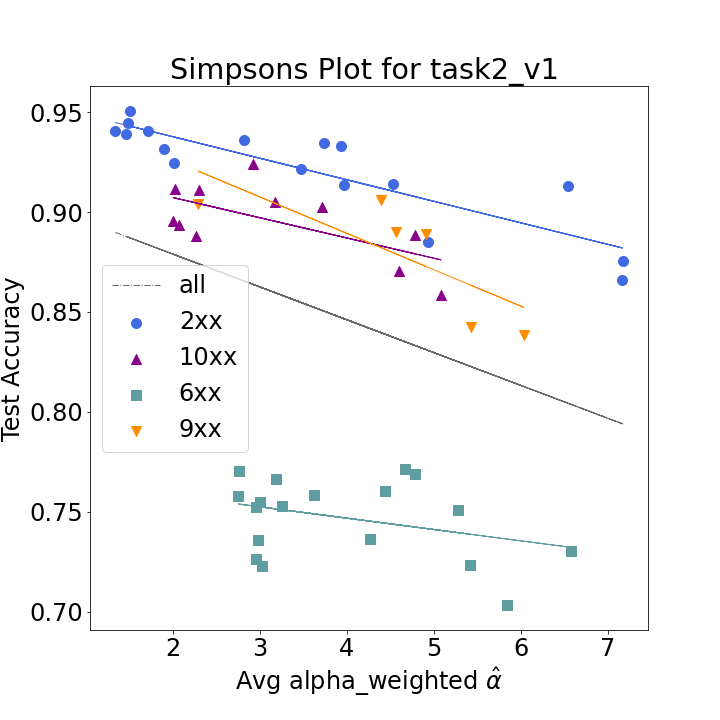}
        \label{fig:simpsonsPlots3-task2}
    }
    \caption{
             Test accuracy versus \ALPHAHAT, for \TASKONE and \TASKTWO, segmented by model sub-group.
             (Similar results are seen for \ALPHASHATTENNORM.)
             Observe that the Simpson's paradox for \TASKTWO models disappears when the data are analyzed with this metric.
    }
    \label{fig:simpsonsPlots3}
\end{figure}

Finally, consider Figure~\ref{fig:simpsonsPlots3}, which shows the test accuracy versus \ALPHAHAT, 
for \TASKONE and \TASKTWO, segmented by model sub-group,
and compare this with 
Figure~\ref{fig:simpsonsPlots1} (for \SPECTRALNORM)
Figure~\ref{fig:simpsonsPlots2} (for \ALPHA).
Recall that \ALPHAHAT may be viewed either as 
a weighted \ALPHA (weighted by the layer $\lambda^{max}_{l}$) or as 
a weighted \SPECTRALNORM (weighted by the layer $\alpha_{l}$). 
As a weighted \ALPHA, the spectral norm weighting corrects for the fact that \ALPHA is scale-invariant, accounting for the variation in the scale of each weight matrix across different layers. 
As a weighted \SPECTRALNORM, the $\alpha$ weighting corrects for the fact that the \SPECTRALNORM is anti-correlated with the test accuracy when varying the regularization hyperparameters.
In either case, this combination ``corrects for'' the Simpson's paradoxes observed in 
Figures~\ref{fig:simpsonsPlots1} and~\ref{fig:simpsonsPlots2}.  
This is likely the explanation for the previously-observed success of this metric at predicting the quality of SOTA DNN models~\cite{MM20a_trends_NatComm}.

\vspace{-3mm}
\section{Conclusion}
\label{sxn:conclusion}
\vspace{-1mm}

A goal of the HT-SR theory is to develop model quality metrics, like \ALPHA and \ALPHAHAT, that can be applied to so-called pre-trained DNN models, by only watching the weights, i.e., without the need for access to training/testing data. 
Previous work demonstrated the utility of the \ALPHAHAT metric when applied to a wide range of large scale and production, pre-trained CV and NLP models~\cite{MM20a_trends_NatComm}. 
Here, we seek to better understand why the \ALPHAHAT metric works so well.

To accomplish this, we evaluated the \ALPHAHAT metric and its subcomponent metrics, 
\ALPHA and \SPECTRALNORM, on a set of publicly-available pre-trained models made available from a recent machine learning contest aimed at understanding causes of good generalization.
To our initial surprise, we identified a clear Simpson's paradox in the data.
Based on our exploration of that, we developed an improved understanding of the complementary roles of \SCALE metrics versus 
\SHAPE metrics in evaluating model quality.
Overall, our analysis explains the success of previously-introduced metrics that combine norm information and shape/correlational information~\cite{MM20a_trends_NatComm} as well as previously-observed peculiarities of norm-based metrics~\cite{JNBx19_fantastic_TR}.
Our results also highlight the need to go beyond one-size-fits-all metrics based on upper bounds from generalization theory to describe the performance of SOTA NNs, as well as the evaluation of models by expensive retraining on test data. 
Test data can be extremely expensive to acquire, and there is a need to evaluate models without (much or any) test data.  %

Based on our findings, we expect that \SPECTRALNORM (and related \SCALE-based metrics) can capture coarse model trends (e.g., large-scale architectural changes, such as width or depth or perhaps adding residual connections and convolutional width) more generally; and that \ALPHA (and related \SHAPE-based metrics) can capture fine-scale model properties (e.g., fine-scale optimizer/solver hyperparameters changes, such as batch size, step scale, etc.) more generally.
Understanding this better is an important future direction raised by our work.
Clearly, our results also have implications beyond generalization and even model training, to problems such as fine-tuning of pre-trained models, improved model engineering, and improved neural architecture search.
(See Appendix~\ref{app:additional-discussion} for some addidional discussion.)
These too are important future directions raised by our work.

\vspace{5mm}
\noindent
\paragraph{Acknowledgements.}
MWM would like to acknowledge ARO, DARPA, IARPA (under contract W911NF20C0035), NSF, and ONR as well as the UC Berkeley BDD project and a gift from Intel for providing partial support of this work.
Our conclusions do not necessarily reflect the position or the policy of our sponsors, and no official endorsement should be inferred.

\bibliographystyle{unsrt}
{\small
%
%%\bibliography{dnns}

\begin{thebibliography}{10}

\bibitem{MM20a_trends_NatComm}
C.~H. Martin, T.~S. Peng, and M.~W. Mahoney.
\newblock Predicting trends in the quality of state-of-the-art neural networks
  without access to training or testing data.
\newblock {\em Nature Communications}, 12(4122):1--13, 2021.

\bibitem{MM18_TR_JMLRversion}
C.~H. Martin and M.~W. Mahoney.
\newblock Implicit self-regularization in deep neural networks: Evidence from
  random matrix theory and implications for learning.
\newblock {\em Journal of Machine Learning Research}, 22(165):1--73, 2021.

\bibitem{MM19_HTSR_ICML}
C.~H. Martin and M.~W. Mahoney.
\newblock Traditional and heavy-tailed self regularization in neural network
  models.
\newblock In {\em Proceedings of the 36th International Conference on Machine
  Learning}, pages 4284--4293, 2019.

\bibitem{MM20_SDM}
C.~H. Martin and M.~W. Mahoney.
\newblock Heavy-tailed {U}niversality predicts trends in test accuracies for
  very large pre-trained deep neural networks.
\newblock In {\em Proceedings of the 20th SIAM International Conference on Data
  Mining}, 2020.

\bibitem{JFYx20_contest_v10}
Y.~Jiang, P.~Foret, S.~Yak, D.~M. Roy, H.~Mobahi, G.~K. Dziugaite, S.~Bengio,
  S.~Gunasekar, I.~Guyon, and B.~Neyshabur.
\newblock {NeurIPS} 2020 competition: Predicting generalization in deep
  learning (version 1.0).
\newblock Technical Report Manuscript, June 28, 2020, 2020.

\bibitem{JFYx20_contest_v11}
Y.~Jiang, P.~Foret, S.~Yak, D.~M. Roy, H.~Mobahi, G.~K. Dziugaite, S.~Bengio,
  S.~Gunasekar, I.~Guyon, and B.~Neyshabur.
\newblock {NeurIPS} 2020 competition: Predicting generalization in deep
  learning (version 1.1).
\newblock Technical Report Preprint: December 16, 2020: arXiv:2012.07976v1,
  2020.

\bibitem{EB01_BOOK}
A.~Engel and C.~P. L.~Van den Broeck.
\newblock {\em Statistical mechanics of learning}.
\newblock Cambridge University Press, New York, NY, USA, 2001.

\bibitem{CSN09_powerlaw}
A.~Clauset, C.~R. Shalizi, and M.~E.~J. Newman.
\newblock Power-law distributions in empirical data.
\newblock {\em SIAM Review}, 51(4):661--703, 2009.

\bibitem{Sim51}
E.~H. Simpson.
\newblock The interpretation of interaction in contingency tables.
\newblock {\em Journal of the Royal Statistical Society. Series B
  (Methodological)}, 13(2):238--241, 1951.

\bibitem{BHO75}
P.~J. Bickel, E.~A. Hammel, and J.~W. O'Connell.
\newblock Sex bias in graduate admissions: Data from {B}erkeley.
\newblock {\em Science}, 187(4175):398--404, 1975.

\bibitem{Rob09}
W.~S. Robinson.
\newblock Ecological correlations and the behavior of individuals.
\newblock {\em International Journal of Epidemiology}, 38:337--341, 2009.

\bibitem{KFWB13}
R.~A. Kievit, W.~E. Frankenhuis, L.~J. Waldorp, and D.~Borsboom.
\newblock Simpson's paradox in psychological science: a practical guide.
\newblock {\em Frontiers in Psychology}, 4(513):1--14, 2013.

\bibitem{JNBx19_fantastic_TR}
Y.~Jiang, B.~Neyshabur, H.~Mobahi, D.~Krishnan, and S.~Bengio.
\newblock Fantastic generalization measures and where to find them.
\newblock Technical Report Preprint: arXiv:1912.02178, 2019.

\bibitem{weightwatcher_package}
{WeightWatcher}, 2018.
\newblock \url{https://pypi.org/project/WeightWatcher/}.

\bibitem{Bar97}
P.~L. Bartlett.
\newblock For valid generalization, the size of the weights is more important
  than the size of the network.
\newblock In {\em Annual Advances in Neural Information Processing Systems 9:
  Proceedings of the 1996 Conference}, pages 134--140, 1997.

\bibitem{NTS15}
B.~Neyshabur, R.~Tomioka, and N.~Srebro.
\newblock Norm-based capacity control in neural networks.
\newblock In {\em Proceedings of the 28th Annual Conference on Learning
  Theory}, pages 1376--1401, 2015.

\bibitem{BFT17_TR}
P.~Bartlett, D.~J. Foster, and M.~Telgarsky.
\newblock Spectrally-normalized margin bounds for neural networks.
\newblock Technical Report Preprint: arXiv:1706.08498, 2017.

\bibitem{AGNZ18_TR}
S.~Arora, R.~Ge, B.~Neyshabur, and Y.~Zhang.
\newblock Stronger generalization bounds for deep nets via a compression
  approach.
\newblock Technical Report Preprint: arXiv:1802.05296, 2018.

\bibitem{ZK16}
L.~Zdeborov{\'a} and F.~Krzakala.
\newblock Statistical physics of inference: thresholds and algorithms.
\newblock {\em Advances in Physics}, 65(5):453--552, 2016.

\bibitem{BKPx20}
Y.~Bahri, J.~Kadmon, J.~Pennington, S.~Schoenholz, J.~Sohl-Dickstein, and
  S.~Ganguli.
\newblock Statistical mechanics of deep learning.
\newblock {\em Annual Review of Condensed Matter Physics}, 11:501--528, 2020.

\bibitem{DDNx20_TR}
G.~K. Dziugaite, A.~Drouin, B.~Neal, N.~Rajkumar, E.~Caballero, L.~Wang,
  I.~Mitliagkas, and D.~M. Roy.
\newblock In search of robust measures of generalization.
\newblock Technical Report Preprint: arXiv:2010.11924, 2020.

\bibitem{LDRC18_TR}
Z.~Liao, T.~Drummond, I.~Reid, and G.~Carneiro.
\newblock Approximate {F}isher information matrix to characterise the training
  of deep neural networks.
\newblock Technical Report Preprint: arXiv:1810.06767, 2018.

\bibitem{TPMx19_TR}
V.~Thomas, F.~Pedregosa, B.~van Merrienboer, P.-A. Mangazol, Y.~Bengio, and
  N.~Le Roux.
\newblock On the interplay between noise and curvature and its effect on
  optimization and generalization.
\newblock Technical Report Preprint: arXiv:1906.07774, 2019.

\bibitem{Pearl09}
J.~Pearl.
\newblock {\em Causality: Models, Reasoning and Inference}.
\newblock Cambridge University Press, 2009.

\bibitem{YTHx22_TR}
Y.~Yang, R.~Theisen, L.~Hodgkinson, J.~E. Gonzalez, K.~Ramchandran, C.~H.
  Martin, and M.~W. Mahoney.
\newblock Evaluating natural language processing models with generalization
  metrics that do not need access to any training or testing data.
\newblock Technical Report Preprint: arXiv:2202.02842, 2022.

\bibitem{newman2005_zipf}
M.~E.~J. Newman.
\newblock Power laws, {P}areto distributions and {Z}ipf's law.
\newblock {\em Contemporary Physics}, 46:323--351, 2005.

\bibitem{BouchaudPotters03}
J.~P. Bouchaud and M.~Potters.
\newblock {\em Theory of Financial Risk and Derivative Pricing: From
  Statistical Physics to Risk Management}.
\newblock Cambridge University Press, 2003.

\bibitem{Sornette07TR}
D.~Sornette.
\newblock Probability distributions in complex systems.
\newblock Technical Report Preprint: arXiv:0707.2194, 2007.

\bibitem{BegTim12}
J.~M. Beggs and N.~Timme.
\newblock Being critical of criticality in the brain.
\newblock {\em Frontiers in Physiology}, 3(163), 2012.

\bibitem{ABP14}
J.~Alstott, E.~Bullmore, and D.~Plenz.
\newblock powerlaw: A python package for analysis of heavy-tailed
  distributions.
\newblock {\em PLoS ONE}, 9(1):e85777, 2014.

\bibitem{MTBx16}
N.~Marshall, N.~M. Timme, N.~Bennett, M.~Ripp, E.~Lautzenhiser, and J.~M.
  Beggs.
\newblock Analysis of power laws, shape collapses, and neural complexity: New
  techniques and {MATLAB} support via the {NCC} toolbox.
\newblock {\em Frontiers in Physiology}, 7(250):1--18, 2005.

\end{thebibliography}
%

}

\appendix
\section{Metrics considered in our analysis}
\label{app:metrics_in_analysis}
\vspace{-2mm}

As mentioned in Section~\ref{sxn:background}, we considered a range of metrics in our analysis. 
See Table~\ref{table:metrics-considered} for a summary of the best performing.
This includes both \emph{average PL metrics}, from HT-SR Theory, and \emph{average log-norm metrics}, from SLT, as well as metrics that combine the two~approaches.
(Other metrics performed worse, exhibited similar qualitative trends, or required access to training and testing data.)

\begin{table}[h] %
\small
\begin{center}
\begin{tabular}{|p{1.5in}|c|c|c|c|c|c|}
\hline
Complexity Metric   &  Average  
                    & Ref. 
                    & \makecell{Scale or \\ Shape? } 
                    & \makecell{Need \\ data? } 
                    & \makecell{Need \\ initial \\ weights? } 
                    & \makecell{Need \\ GPUs? } \\
\hline
 \SPECTRALNORM      &  $ \AVGLOGSPECTRALNORM $ & (\cite{JNBx19_fantastic_TR}) & Scale  &  No &  No &  No \\
 \FROBENIUSNORM     &  $ \AVGLOGNORM $         & (\cite{JNBx19_fantastic_TR}) & Scale &  No &  No &  No \\
\hline
 \ALPHA             &  $ \AVGALPHA $           & (this paper)                 & Shape &  No &  No &  No \\
 \QUALITYOFALPHAFIT &  $ D_{KS} $              & (this paper)                 & Shape &  No &  No &  No \\
\hline
 \ALPHAHAT          &  $ \ALPHAHATEQN $        & (\cite{MM20a_trends_NatComm})& Both  &  No &  No &  No \\
 \ALPHASHATTENNORM  &  $ \AVGLOGSHATTENNORM $  & (\cite{MM20a_trends_NatComm})& Both  &  No &  No &  No \\
\hline
\end{tabular}
\end{center}
\caption{Overview of model quality metrics.
         Based on our initial analysis of Contest models, we propose and evaluate \ALPHA and \QUALITYOFALPHAFIT.  %
         The metrics in this table do not need access to training/testing data; they do not need information such as the initial weight distribution; and they do not require training/retraining (and thus access to GPUs).  %
}
\label{table:metrics-considered}
\end{table}

\paragraph{Average Power-Law (PL) metrics.}

Given a (pre-)trained DNN with $L$ layer weight matrices $\mathbf{W}$, PL metrics are computed by fitting the ESD of the correlation matrix $\mathbf{X}=\mathbf{W}^{T}\mathbf{W}$ of each layer to a PL, and then averaging over all layers.%
\footnote{For Conv2D$(k,k,N,M)$ layers, we extract $N=k \times k$ matrices of shape $N \times M$; typically, $k=1$ or $3$.} 
In more detail, each layer ESD is fit to a PL of the form:
$$
\rho(\lambda)\sim\lambda^{-\alpha},\;\; x_{min} \le \lambda \le x_{max}.
$$
The fitting procedure selects the optimal PL exponent $\alpha$, and the adjustable parameter $x_{min}$.  
The PL exponent $\alpha$ characterizes the tail of the ESD; it measures what may be interpreted as the ``shape'' of most important part of the spectrum. 

\begin{itemize}[leftmargin=*,noitemsep,nolistsep]
\item 
\ALPHA: $ \AVGALPHA =\frac{1}{L}\sum_{l}\alpha_{l} $.
This is a simple average of fitted $\alpha$ over all layers $L$.
From the perspective of statistical mechanics, \ALPHA quantifies  amount of correlation in the layer weight matrices.%
\footnote{Prior work \cite{MM18_TR_JMLRversion} has argued that NNs resemble the strongly correlated systems, e.g., in electronic structure theory and statistical mechanics, which was the origin of early work in heavy-tailed random matrix theory \cite{BouchaudPotters03}.} 
From the perspective of statistics, \ALPHA may be viewed a shape parameter.

\item 
\QUALITYOFALPHAFIT: $ \AVGALPHADISTANCE $.
For the given set of parameters $(\alpha, x_{min}, x_{max})$, the quality of the PL fit can be measured by the Kolmogorov-Smirnov (KS) distance ($D_{KS}$) between the empirical and theoretical distributions.  The smaller $D_{KS}$ is, the better the fit.
\end{itemize}

\noindent
We describe the PL fitting procedure in more detail (in Section~\ref{sxn:scale_shape_parameters-overview}), and we give examples of both high and low quality fits (in Appendix~\ref{app:fitting_tpls}).
To our knowledge, we are the first to use \ALPHA and \QUALITYOFALPHAFIT to gauge model~complexity.%
\footnote{Prior work~\cite{MM20a_trends_NatComm} used a weighted version of this metric (\ALPHAHAT).  There, PL exponents $\alpha$ were computed, but they were not evaluated as a measure of test accuracy, and they were not shown to correlate with variations in solver hyperparameters (such as batch size, dropout, weight decay, etc.), as we do for the first time here.  }

\paragraph{Average Log-Norm metrics.}

Log-Norm metrics are related to \emph{product-norm measures} of the model complexity $\mathcal{C}$.  
Given a (pre-)trained DNN with $L$ layers, and layer weight matrices $\mathbf{W}_{l}$, we define the $\mathcal{C}$ as
the product over a norm of the layer weight matrices
\begin{equation}
\mathcal{C}:=\Vert\mathbf{W}_{1}\Vert\times\Vert\mathbf{W}_{2}\Vert\times\cdots\Vert\mathbf{W}_{L}\Vert  ,
\end{equation}
where $\Vert\mathbf{W}_{l}\Vert$ denotes some arbitrary matrix norm for layer $l$.%
\footnote{We drop the layer subscript $l$ when it is clear from the context.}
If we take the logarithm of both side, we can express this complexity as an average over all layers ($L$), in log-units
\begin{equation}
\log_{10}\mathcal{C}\sim\langle\log_{10}\Vert\mathbf{W}\Vert\rangle:=\dfrac{1}{L}\sum_{l=1}^{N}\log_{10}\Vert\mathbf{W}_{l}\Vert  .
\end{equation}
Metrics of this form provide a measure of the ``scale'' (in log units) of a model.

\begin{itemize}[leftmargin=*,noitemsep,nolistsep]
\item
\SPECTRALNORM: $ \AVGLOGSPECTRALNORM $.
This is the (average) of the log of the layer Spectral norms.
The layer Spectral Norm is just maximum eigenvalue $\lambda^{max}$ of the correlation matrix $\mathbf{X}$.%
\footnote{Prior work has shown that using norm-based metrics in log-scale tends to be superior to working with them in non-log scale~\cite{MM20a_trends_NatComm}.  Even when taking averages, however, norm-based metrics, unlike \ALPHA, are \emph{not} scale~invariant.}

\item 
\FROBENIUSNORM: $ \AVGLOGNORM $. 
This is the average of the log of the Frobenius norm for each layer weight matrix. 
It is included for completeness. 
\end{itemize}

\noindent
Importantly, these metrics take a layer average, and not a sum, 
since otherwise these metrics will trivially depend on the depth $L$ of the network.

\paragraph{Combining PL and Log-Norm metrics.}

Previous work considered metrics that combine ``scale'' and ``shape'' ideas~\cite{MM20_SDM,MM20a_trends_NatComm}, two of which we consider here. 

\begin{itemize}[leftmargin=*,noitemsep,nolistsep]

\item 
\ALPHAHAT: $ \ALPHAHATEQN=\frac{1}{L}\sum_{l}\alpha_{l}\log_{10}\lambda^{max}_{l} $.
This has been used previously as a complexity measure for a large number of pre-trained NNs of varying depth and hyperparameters~\cite{MM20_SDM,MM20a_trends_NatComm}. 
\ALPHAHAT may be viewed in one of two complementary ways: 
either as a weighted \ALPHA (weighted by the layer $\lambda^{max}_{l}$); 
or, equivalently, as a weighted \SPECTRALNORM (weighted by the layer $\alpha_{l}$). 
As a weighted \ALPHA, the spectral norm weighting corrects for the fact that \ALPHA is scale-invariant, accounting for the variation in the scale of each weight matrix across different layers. 
As a weighted \SPECTRALNORM, the $\alpha$ weighting corrects for the fact that the \SPECTRALNORM is anti-correlated with the test accuracy when varying the regularization hyperparameters.

\item 
\ALPHASHATTENNORM: 
$\AVGLOGSHATTENNORM = \langle\log_{10}\Vert\mathbf{X}\Vert^{\alpha}_{\alpha}\rangle$.
This has been used previously~\cite{MM20_SDM}, and it is the average of the Log of the standard Shatten norm, defined with the value of $2\alpha$ from the PL fit of the ESD of each layer weight matrix $\mathbf{W}$. 
The \ALPHAHAT metric approximates the \ALPHASHATTENNORM under certain conditions~\cite{MM20_SDM}.

\end{itemize}

\noindent
Table~\ref{table:metrics-considered} provides a summary of the metrics we considered.
There are many other metrics that we examined but that we do not describe here:
the path norm and Fisher-Rao metcics~\cite{JFYx20_contest_v10,JFYx20_contest_v11} are more expensive, perform worse, and/or don't add new insight;
the Jacobian Norm
is also too expensive to compute; 
and other metrics described in the Contest~\cite{JFYx20_contest_v10,JFYx20_contest_v11} either were uninteresting (e.g., the sum---not average---of layer norms, which is a proxy for depth) or performed very poorly.

\section{Models considered in our analysis}
\label{app:models_in_analysis}
\vspace{-2mm}

See Table~\ref{table:models-from-contest} for a summary of the models we considered in our analysis.
These are described in more detail in Section~\ref{sxn:background}.
See also~\cite{JFYx20_contest_v10,JFYx20_contest_v11} for more details.

\begin{table}[h]  %
\small
\begin{center}
\begin{tabular}{|p{0.75in}|c|c|c|c|c|c|c|c|c|}
\hline
 Series                 &\#    & L & Batch Size & Dropout & Weight Decay & Conv Width & Dense & (k) \\
\hline
 \TASKONE               &  0xx & 4 & 8, 32, 512 & 0.0, 0.5 & 0.0, 0.001 & 256, 512 & 1 & 1 \\
 ``task1\_v4''          &  1xx & 5 & 8, 32, 512 & 0.0, 0.5 & 0.001 & 256, 512 & 2 & 1\\
 (VGG-like)             &  2xx & 5 & 8, 32, 512 & 0.0, 0.5 & 0.0 & 256, 512 & 2 & 1\\
                        &  5xx & 8 & 8, 32, 512 & 0.0, 0.5 & 0.001 & 256, 512 & 1 & 3\\
                        &  6xx & 8 & 8, 32, 512 & 0.0, 0.5 & 0.0 & 256, 512 & 2 & 3\\
                        &  7xx & 9 & 8, 32, 512 & 0.0, 0.5 & 0.0 & 256, 512 & 2 & 3\\
\hline
 \TASKTWO               &  2xx & 13 & 32, 512, 1024 & 0.0, 0.25, 0.5 & 0.0, 0.001 & 512 & -  & - \\
 ``task2\_v1''          &  6xx & 7 & 32, 512, 1024 & 0.0, 0.25, 0.5 & 0.0, 0.001 & 512 & - & - \\
 (Network-              &  9xx & 10 & 1024 & 0.0, 0.25, 0.5 & 0.0, 0.001 & 512 & - & - \\
  in-Network)           & 10xx & 10 & 32, 512, 1024 & 0.0, 0.25, 0.5 & 0.0, 0.001 & 512 & - & -\\
\hline
\end{tabular}
\end{center}
\caption{Overview of models we considered in each task and sub-group, including variations in depth (L), regularization hyperparameters (Batch Size, amount of Dropout, and Weight Decay), and architectural changes (Width of selected Convolutional Layers, number of selected Dense layers, and kernel-size $(k=1,3)$ for selected Convolutional layers).  See~\cite{JFYx20_contest_v10,JFYx20_contest_v11} for complete~details. 
        }
\label{table:models-from-contest}
\end{table}

\section{More on determining shape parameters (fitting ESDs to PLs),
from Section~\ref{sxn:scale_shape_parameters-fitting}
}
\label{app:fitting_tpls}
\vspace{-2mm}

Fitting data to PLs is very finicky~\cite{newman2005_zipf,Sornette07TR,CSN09_powerlaw,BegTim12,ABP14,MTBx16}.
We have found it best to proceed with a combination of visual inspection and analysis with the \texttt{WeightWatcher} tool~\cite{weightwatcher_package}.

\paragraph{Visual inspection of ESDs.}

\begin{figure}[t]  %
    \centering
    \subfigure[ESD (log-log plot)]{
        \includegraphics[width=3.7cm]{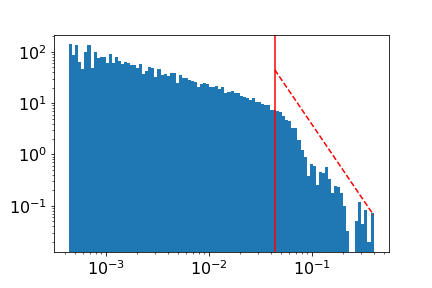}
        \label{fig:good-vs-bad-pl-fits-21}
    }
    \subfigure[ESD (lin-lin plot)]{
        \includegraphics[width=3.7cm]{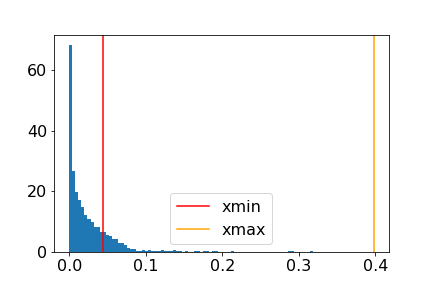}
        \label{fig:good-vs-bad-pl-fits-22}
    }
    \subfigure[ESD (log-lin plot)]{
        \includegraphics[width=3.7cm]{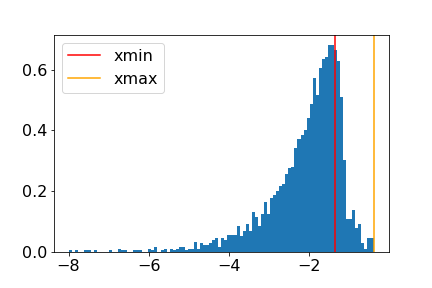}
        \label{fig:good-vs-bad-pl-fits-23}
    }
    \subfigure[PL fit quality vs $x_{min}$]{
        \includegraphics[width=3.7cm]{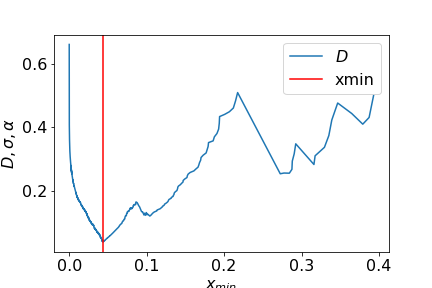}
        \label{fig:good-vs-bad-pl-fits-24}
    } \\
    \subfigure[ESD (log-log plot)]{
        \includegraphics[width=3.7cm]{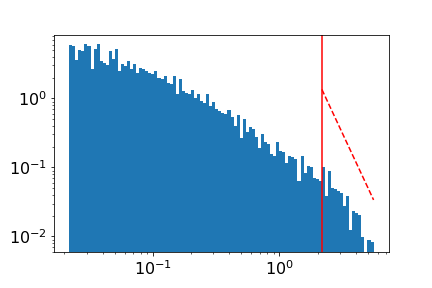}
        \label{fig:good-vs-bad-pl-fits-31}
    }
    \subfigure[ESD (lin-lin plot)]{
        \includegraphics[width=3.7cm]{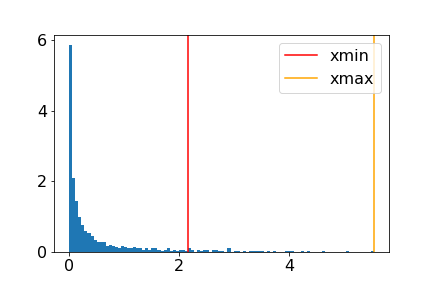}
        \label{fig:good-vs-bad-pl-fits-32}
    }
    \subfigure[ESD (log-lin plot)]{
        \includegraphics[width=3.7cm]{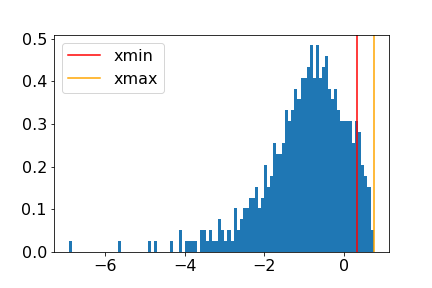}
        \label{fig:good-vs-bad-pl-fits-33}
    }
    \subfigure[PL fit quality vs $x_{min}$]{
        \includegraphics[width=3.7cm]{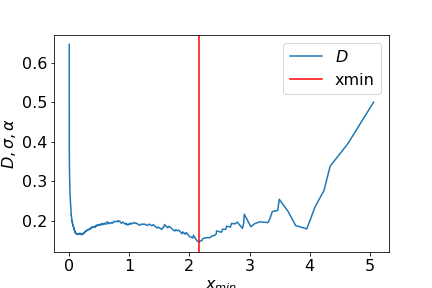}
        \label{fig:good-vs-bad-pl-fits-34}
    } \\
    \caption{Illustration of the role of the ESD shape in determining the PL parameter $\alpha$ in the PL fit.  
             Rows correspond to different layers in different models. 
             Columns correspond to viewing the same ESD in different ways---on a log-log plot (first column), a linear-linear plot (second column), and a log-linear plot (third)---and the KS Distance $D$ of the PL fit as a function of $x_{min}$ (final column).
             Row 1: %
             \TASKONE, 
             model 152, layer 1.  
             Row 2: %
             \TASKTWO, 
             model 1006, layer 10.
            }
    \label{fig:good-vs-bad-pl-fits}
\end{figure}

The following advice, taken directly from Sornette~\cite{Sornette07TR}, is particularly helpful for visual inspection of ESDs:
``we recommend a preliminary visual exploration by plotting the survival and density distributions in (i) linear-linear coordinates, (ii) log-linear coordinates (linear abscissa and logarithmic ordinate) and (iii) log-log coordinates (log-arithmic abscissa and logarithmic ordinate).  
The visual comparison between these three plots provides a fast and intuitive view of the nature of the data.
\begin{itemize}
\item
A power law distribution will appear as a convex curve in the linear-linear and log-linear plots and as a straight line in the log-log plot.
\item
A Gaussian distribution will appear as a bell-shaped curve in the linear-linear plot, as an inverted parabola in the log-linear plot and as strongly concave sharply falling curve in the log-log plot.
\item
An exponential distribution will appear as a convex curve in the linear-linear plot, as a straight line in the log-linear plot and as a concave curve in the log-logp lot.
\end{itemize}
Having in mind the shape of these three reference distributions in these three representations provides fast and useful reference points to classify the unknown distribution under study.'' 
It is helpful to examine ESDs such as those in 
Figure~\ref{fig:good-vs-bad-pl-fits-ideal} (in Sectionn~\ref{sxn:scale_shape_parameters-fitting}) and Figure~\ref{fig:good-vs-bad-pl-fits} (here) in light of these~comments.

Of course, such a visual inspection is just a first step to a more detailed analysis, since by itself visual analysis of HT data can be misleading.
Sornette~\cite{Sornette07TR} goes on to say:
``While we recommend a first visual inspection,
it is only a first indication, not a proof.  
It is a necessary step to convince oneself (and the reviewers and journal editors) but certainly not a sufficient condition.  
It is a standard rule of thumb that a power law scaling is thought to be meaningful if it holds over at least two to three decades on both axes and is bracketed by deviations on both sides whose origins can be understood (for instance, due to insufficient samplingand/or finite-size effects).''
It is for these reasons that understanding the behavior of the ESDs near $x_{min}$ and $x_{max}$ (of Eqn.~(\ref{eqn:tpl_eqn})) is so important.

\paragraph{Fitting to non-ideal ESDs.}

In Figure~\ref{fig:good-vs-bad-pl-fits-ideal}, we illustrated how PL fits of ESDs perform on a nearly ``ideal'' example.
Here, we discuss how it performs on less-than-ideal examples (that occurred in the Contest data).
See Figure~\ref{fig:good-vs-bad-pl-fits}.

In Figure~\ref{fig:good-vs-bad-pl-fits}, 
we see several examples of layers that are less well-fit by a PL.
In these cases, we see that 
the linear-linear plots are non-informative; 
the log-linear plots show that the distributions have a strong left-ward bias, 
indicating a tail of increasingly small eigenvalues; and 
there is not a broad range of large eigenvalues on the right of the distribution.
This relative paucity of large eigenvalues is seen on the log-linear plots by a (more or less aggressive) truncation in probability mass for larger eigenvalues, and on the log-log plots by a steeper downward slope on the right of the ESD.
(In these cases, compare with Figure~\ref{fig:good-vs-bad-pl-fits-ideal}.)
The first row  %
(\TASKONE, model 152, layer 1)
illustrates 
an incompletely-developed right tail 
(alternatively a ``bulk-plus-spikes'' model may be a more appropriate fit than a PL fit), 
meaning that the spectral norm is somewhat smaller and that the fitting procedure has difficulty choosing $x_{min}$ near the peak of the distribution.
The second row  %
(\TASKTWO, model 1006, layer 10)
illustrates 
an aggressively-shortened (effectively truncated, i.e., not even spikes) right tail, which leads to a much smaller spectral norm ($x_{max}$) and thus a much larger $x_{min}$ (since there is such a small range over which a linear fit is appropriate), as well as a broad range of (large) $x_{min}$ values over which low-quality fits are obtained. 
In each of these of these cases, the KS distance plots have less of a well-defined minimum as a function of $x_{min}$.

\paragraph{Additional discussion.}

In simple cases, scale and shape parameters do capture similar information.
For models whose ESDs are very well-approximated by a Marchenko-Pastur (MP) distribution or a MP bulk-plus-spike distribution (\textsc{Random-like} and \textsc{Bulk+Spikes} phases from \cite{MM18_TR_JMLRversion}), visual inspection of ESD plots 
often yields insight, and there is a strong correspondence between norm-based scale metrics and random MP-based shape metrics.
Similarly, for models whose ESDs are very well-approximated by PL distributions (\textsc{Heavy-Tailed} phase from \cite{MM18_TR_JMLRversion}, where the $x_{max} = \lambda_{max}$ truncation is not signigicant~\cite{MM18_TR_JMLRversion} to the fit), there is a strong correspondence between norm-based scale metrics and PL-based shape metrics, in the sense that smaller $\alpha$ values correspond closely to larger $\lambda_{max}$ values. 
For realistic models, however, and for the models from Table~\ref{table:models-from-contest}, these two metrics can be very different and can reveal very different information.

The broad range of behavior seen in the Contest data arises since ESDs look like ones in the 
rows of Figure~\ref{fig:good-vs-bad-pl-fits} (rather than the more ``ideal'' case shown in Figure~\ref{fig:good-vs-bad-pl-fits-ideal}) where the linear fit to the ESD on log-scale is not very good.
For models where \ALPHA against \SPECTRALNORM behave more similarly, the ESDs look more like 
Figure~\ref{fig:good-vs-bad-pl-fits-ideal}.
For the Contest models shown in Figure~\ref{fig:good-vs-bad-pl-fits} (and others),
for a simple (non-truncated, long tail) PL distribution, 
smaller exponents $\alpha$ correspond to a longer tail, which corresponds to a larger $\lambda_{max}$.
However, since the tail of the ESDs are typically best fit by a PL---often with exponents $\alpha$ much larger than expected, since many of the models from Table~\ref{table:models-from-contest} are of lower quality---the situation is not so simple.
This simple connection only holds for a few model sub-groups.
Generally speaking, these scale and shape metrics (\SPECTRALNORM and \ALPHA) characterize the model ESDs differently and capture different properties of the~models.

\section{Illustrative examples: comparing scale versus shape parameters}
\label{app:illustrative_scale_shape}
\vspace{-2mm}

\begin{figure}[t]  %
    \centering
    \subfigure[\TASKONE, $2xx$]{
        \includegraphics[width=3.7cm]{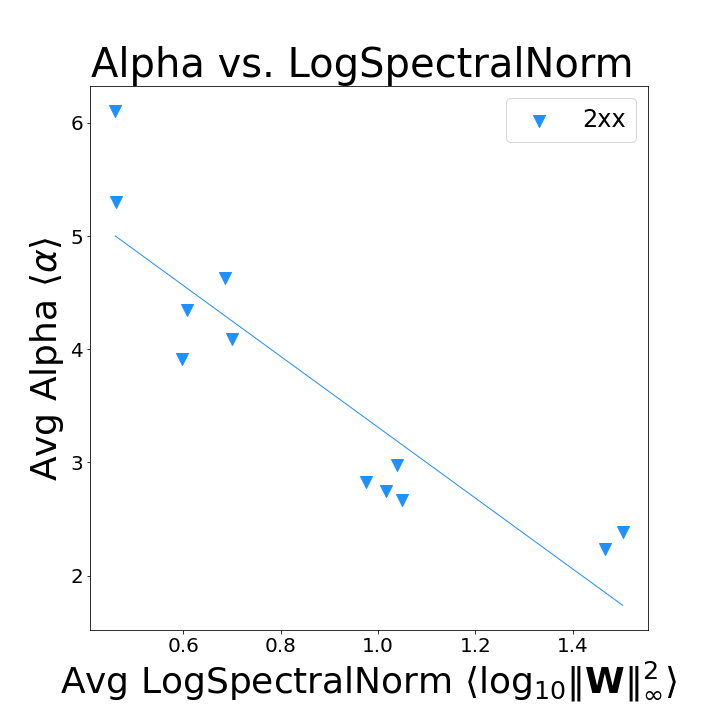}
        \label{fig:alpha-vs-spnorm-A}
    }
    \subfigure[\TASKONE, $5xx$]{
        \includegraphics[width=3.7cm]{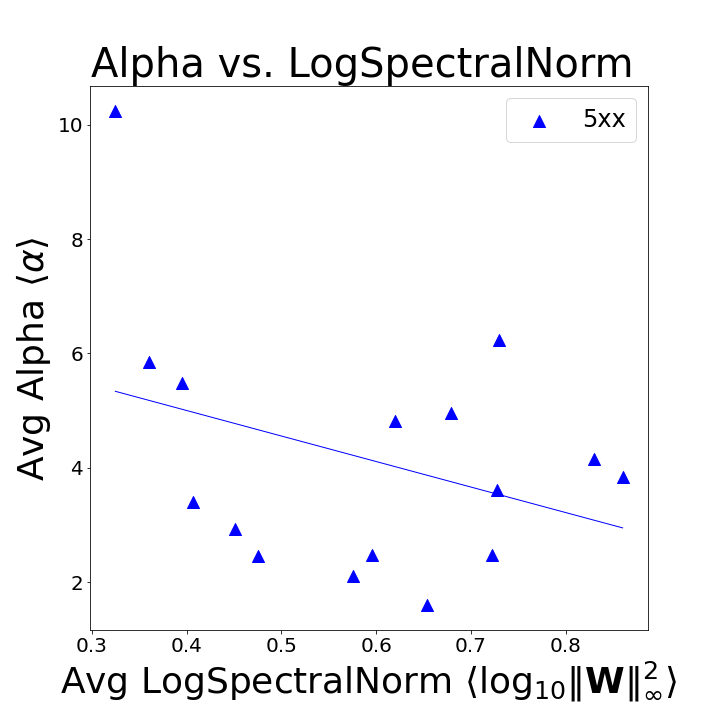}
        \label{fig:alpha-vs-spnorm-B}
    }
    \subfigure[\TASKTWO, $2xx$]{
        \includegraphics[width=3.7cm]{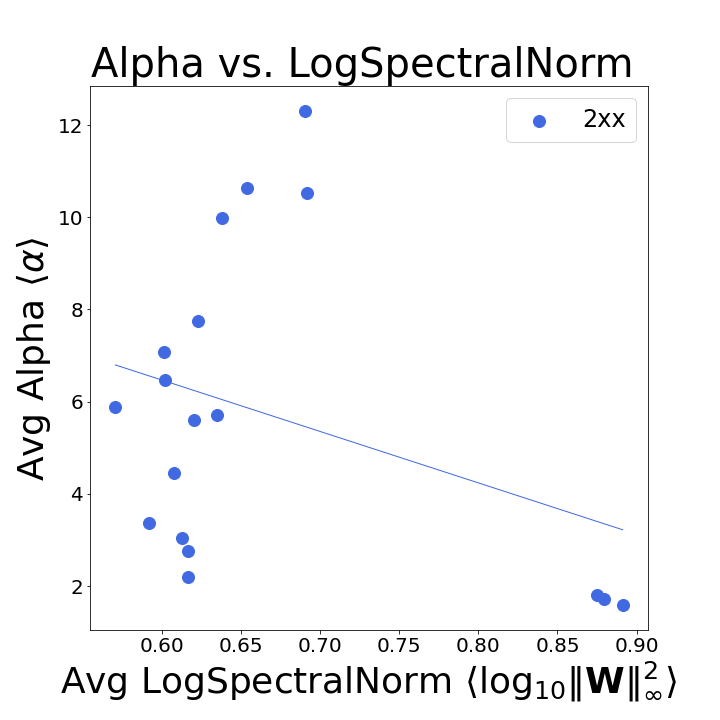}
        \label{fig:alpha-vs-spnorm-C}
    }
    \subfigure[\TASKTWO, $10xx$]{
        \includegraphics[width=3.7cm]{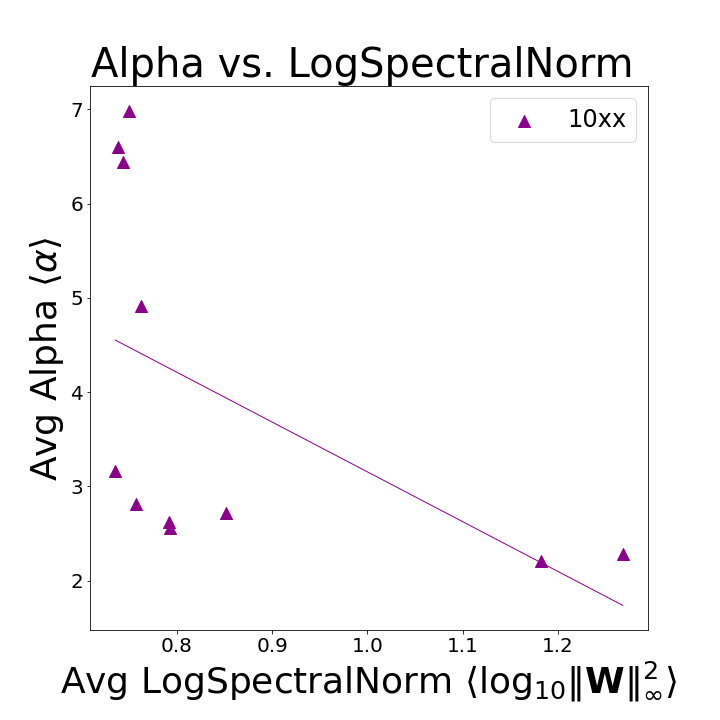}
        \label{fig:alpha-vs-spnorm-D}
    }
    \caption{Comparison of \ALPHA and \SPECTRALNORM for selected model sub-groups from both tasks.  
             Lines (simply to guide the eye) shows a linear regression on the data.
            }
    \label{fig:alpha-vs-spnorm}
\end{figure}

We saw 
in Figure~\ref{fig:alpha-versus-spectral}
 a comparison of the \ALPHA and \SPECTRALNORM metrics, for \TASKONE and \TASKTWO models.
To get more detailed insight,
Figure~\ref{fig:alpha-vs-spnorm} plots \ALPHA versus \SPECTRALNORM for two illustrative pairs of examples, from each of \TASKONE and \TASKTWO.
Consider, as a baseline example, \TASKONE, model sub-group $2xx$, in Figure~\ref{fig:alpha-vs-spnorm-A}.
Here, the two metrics are strongly anti-correlated, with linear correlation metric $R^2=0.803$ and Kendall-$\tau$ rank correlation metric $\tau=0.788$.
In contrast, for \TASKONE, model sub-group $5xx$, shown in \ref{fig:alpha-vs-spnorm-B}, the two metrics are (at best) only weakly correlated, with $R^2=0.124$ and $\tau=0.177$.  
This is typical; some model sub-groups exhibit large rank and/or linear correlations, others virutally none at all.
For \TASKTWO, model sub-group $2xx$, versus \TASKTWO, model sub-group $10xx$, shown in \ref{fig:alpha-vs-spnorm-C} and \ref{fig:alpha-vs-spnorm-D}, we see an example where the two plots look similar visually, and both have small(ish) linear correlation $R^2$, but they have very different Kendall-$\tau$ rank correlation metrics. 
\section{Additional details on changing architectures versus changing solver hyperparameters}
\label{app:details-simpsons-changing}
\vspace{-2mm}

\begin{table}[h] %
\small
\begin{center}

\begin{tabular}{|p{1.05in}|c|c|c|c|c|c|c|c|}
\hline
& \multicolumn{2}{|c|}{ \SPECTRALNORM      } 
& \multicolumn{2}{|c|}{ \FROBENIUSNORM     } 
& \multicolumn{2}{|c|}{ \ALPHA             } 
\\
\hline
               &  $R^2$ & Kendall-$\tau$ &  $R^2$ & Kendall-$\tau$ &  $R^2$ & Kendall-$\tau$  \\
\hline
\TASKONE - 0xx &  0.34  &  0.41  &  0.00  &  -0.03  & \textbf{ 0.55 } & \textbf{ -0.55 }\\
\hline
\TASKONE - 1xx &  0.68  & \textbf{ 0.73 } &  0.29  &  -0.30  & \textbf{ 0.88 } &  -0.48 \\
\hline
\TASKONE - 2xx &  0.59  &  0.73  &  0.42  &  0.55  & \textbf{ 0.69 } &  -0.52 \\
\hline
\TASKONE - 5xx & \textbf{ 0.58 } & \textbf{ 0.60 } &  0.19  &  -0.25  &  0.09  &  -0.10 \\
\hline
\TASKONE - 6xx &  0.22  & \textbf{ 0.59 } &  0.05  &  -0.10  & \textbf{ 0.67 } &  -0.32 \\
\hline
\TASKONE - 7xx &  0.58  & \textbf{ 0.61 } &  0.38  &  0.58  &  0.45  &  -0.51 \\
\hline
\TASKONE - AVG &  0.50  & \textbf{ 0.61 } &  0.22  &  0.07  & \textbf{ 0.55 } &  -0.41 \\
\hline
\hline
\TASKTWO - 2xx &  0.05  &  0.01  &  0.67  & \textbf{ -0.84 } &  0.69  &  -0.74 \\
\hline
\TASKTWO - 10xx &  0.32  &  0.60  &  0.60  &  -0.53  &  0.62  & \textbf{ -0.67 }\\
\hline
\TASKTWO - 6xx &  0.21  & \textbf{ 0.36 } &  0.01  &  -0.05  & \textbf{ 0.36 } &  -0.25 \\
\hline
\TASKTWO - 9xx &  0.17  &  0.47  &  0.65  & \textbf{ -0.87 } &  0.64  &  -0.87 \\
\hline
\TASKTWO - AVG &  0.19  &  0.36  &  0.48  &  -0.57  &  0.58  &  -0.63 \\
\hline
\end{tabular}

\end{center}
\caption{Model quality for different metrics (all those mentioned in Table~\ref{table:metrics-considered}), for \TASKONE and \TASKTWO, both overall and by model sub-group.
         (This table is part 1 of 2; see also Table~\ref{table:quality_table_2}.)
         Bar plots for \ALPHA and \SPECTRALNORM data from here are shown in Figure~\ref{fig:alpha-vs-spnorm-test-accs}.
}
\label{table:quality_table_1}
\end{table}

See Table~\ref{table:quality_table_1} and Table~\ref{table:quality_table_2} for additional details on changing architectures versus changing solver hyperparameters, from Section~\ref{sxn:simpsons_paradox-details}.

\begin{table}[h] %
\small
\begin{center}

\begin{tabular}{|p{1.05in}|c|c|c|c|c|c|c|c|}
\hline
& \multicolumn{2}{|c|}{ \ALPHAHAT      } 
& \multicolumn{2}{|c|}{ \ALPHASHATTENNORM     } 
& \multicolumn{2}{|c|}{ \QUALITYOFALPHAFIT             } 
\\
\hline
               &  $R^2$ & Kendall-$\tau$ &  $R^2$ & Kendall-$\tau$ &  $R^2$ & Kendall-$\tau$  \\
\hline
\TASKONE - 0xx &  0.00  &  -0.06  &  0.00  &  -0.02  &  0.02  &  -0.10 \\
\hline
\TASKONE - 1xx &  0.01  &  0.00  &  0.00  &  -0.03  &  0.00  &  0.15 \\
\hline
\TASKONE - 2xx &  0.49  &  0.70  &  0.48  &  0.61  &  0.67  & \textbf{ -0.82 }\\
\hline
\TASKONE - 5xx &  0.01  &  0.10  &  0.00  &  0.08  &  0.04  &  0.18 \\
\hline
\TASKONE - 6xx &  0.00  &  -0.04  &  0.01  &  -0.03  &  0.10  &  -0.13 \\
\hline
\TASKONE - 7xx &  0.04  &  -0.06  &  0.07  &  -0.03  & \textbf{ 0.60 } &  -0.58 \\
\hline
\TASKONE - AVG &  0.09  &  0.11  &  0.09  &  0.10  &  0.24  &  -0.22 \\
\hline
\hline
\TASKTWO - 2xx &  0.77  &  -0.75  &  0.62  &  -0.73  & \textbf{ 0.89 } &  -0.83 \\
\hline
\TASKTWO - 10xx &  0.41  &  -0.38  &  0.46  &  -0.42  & \textbf{ 0.70 } &  -0.67 \\
\hline
\TASKTWO - 6xx &  0.12  &  -0.14  &  0.12  &  -0.08  &  0.35  &  -0.25 \\
\hline
\TASKTWO - 9xx &  0.59  &  -0.87  &  0.65  &  -0.87  & \textbf{ 0.91 } &  -0.87 \\
\hline
\TASKTWO - AVG &  0.47  &  -0.53  &  0.46  &  -0.53  & \textbf{ 0.71 } & \textbf{ -0.66 }\\
\hline
\end{tabular}

\end{center}
\caption{Model quality for different metrics (all those mentioned in Table~\ref{table:metrics-considered}), for \TASKONE and \TASKTWO, both overall and by model sub-group.
         (This table is part 2 of 2; see also Table~\ref{table:quality_table_1}.)
         Bar plots for \ALPHA and \SPECTRALNORM data from here are shown in Figure~\ref{fig:alpha-vs-spnorm-test-accs}.
}
\label{table:quality_table_2}
\end{table}

\section{Additional discussion}
\label{app:additional-discussion}
\vspace{-2mm}

We conclude with a few more general thoughts on our results.

As is well known, trying to extract causality from correlation is difficult---precisely since there may be Simpson's-like paradoxes present in the data, depending on how the data are partitioned.
When confronted with a Simpson's paradox, one is tempted to ask whether the marginal associations or partial associations are correct.
Often, the answer is that both are correct, depending on what is of interest.
In particular, simple statistical analysis does not provide any guidance as to causal relationships and whether the marginal association or the partial association is the spurious relationship.
For that reason, rather than trying to extract causality, we took a different approach: we looked for a Simpson's paradox; and when we found it, we tried to interpret it in terms of scale versus shape metrics from SLT and HT-SR theory.

One might wonder ``why'' \SPECTRALNORM and \ALPHA perform as well as they do, when restricted to changes in the model depth and/or solver hyperparameters, respectively.
Establishing such a causal explanation, of course, requires going beyond the data at hand and requires some sort of counterfactual analysis.
This is beyond the scope of this paper. 
A plausible hypothesis, however, is the following.
Since upper bounds from SLT suggest that models with smaller norms have less capacity, these norms are used (either explicitly as a regularizer, or implicitly by adjusting large matrix elements/columns/rows) during the training process, in particular when one varies coarse model parameters such as depth.
On the other hand, coming from HT-SR theory, \ALPHA is not used explicitly or implicitly during the training process.  
Instead, the training process extracts correlations over many size scales from the data, and it is these correlations that are captured by smaller \ALPHA values, consistent with HT-SR theory and practice~\cite{MM18_TR_JMLRversion,MM19_HTSR_ICML}.
This hypothesis is consistent with ``why'' fitted PL metrics from HT-SR theory---in particular the fitted \ALPHAHAT metric--perform so well, both for the models considered in this contest, as well as for a much wider range of publicly-available SOTA DNN Models~\cite{MM20a_trends_NatComm}.
Testing this hypothesis is an important question raised by our results.

\end{document}